\documentclass[12pt,letterpaper]{article}

\usepackage[top=01in,left=1in,footskip=1in,marginparwidth=1in]{geometry}

\usepackage[utf8]{inputenc}

\usepackage{hyperref}
\usepackage{mathrsfs}
\usepackage{booktabs} % For better table formatting
\usepackage{array}    % For better table column formatting
\usepackage{float}
\usepackage{graphicx}
\usepackage{subcaption}
\usepackage{multirow}
\usepackage{adjustbox}
\usepackage{caption}
\usepackage{url}
\usepackage{placeins}
\usepackage{appendix}
\usepackage{makecell}
\usepackage{xcolor}
\usepackage{soul}
\usepackage{bm}
\usepackage{amssymb}
\usepackage{amsmath}
\usepackage{cleveref}
\sethlcolor{blue!30}

\title{Beyond the Kolmogorov Barrier: A Learnable Weighted Hybrid Autoencoder for Model Order Reduction}

\author{%%%% Author details
Nithin Somasekharan$^{1}$, Shaowu Pan$^{1}$\footnote{$^{1}$Department of Mechanical, Aerospace, and Nuclear Engineering, Rensselaer Polytechnic Institute, 110 8th St, Troy, NY, USA, 12180}}

\begin{document}

\maketitle
%%%% Article title to be placed here

%%%% Subject entries to be placed here %%%%
%\subject{fluid mechanics, computational mathematics, artificial intelligence}

%%%% Keyword entries to be placed here %%%%
%\keywords{Reduced-order modeling, Dimensionality Reduction, POD, Deep Learning, Surrogate Modeling}

%%%% Insert corresponding author and its email address}
%\corres{Shaowu Pan\\
%\email{pans2@rpi.edu}}

%%%% Abstract text to be placed here %%%%%%%%%%%%
\begin{abstract}
Representation learning for high-dimensional, complex physical systems aims to identify a low-dimensional intrinsic latent space, which is crucial for reduced-order modeling and modal analysis. To overcome the well-known Kolmogorov barrier, deep autoencoders (AEs) have been introduced in recent years, but they often suffer from poor convergence behavior as the rank of the latent space increases. To address this issue, we propose the \emph{learnable weighted hybrid autoencoder}, a hybrid approach that combines the strengths of singular value decomposition (SVD) with deep autoencoders through a learnable weighted framework. We find that the introduction of learnable weighting parameters is essential --- without them, the resulting model would either collapse into a standard POD or fail to exhibit the desired convergence behavior. Interestingly, we empirically find that our trained model has a sharpness thousands of times smaller compared to other models, \textcolor{black}{which in turn enhances its robustness to input noise}. Our experiments on classical chaotic PDE systems, including the 1D Kuramoto-Sivashinsky and forced isotropic turbulence datasets, demonstrate that our approach significantly improves generalization performance compared to several competing \textcolor{black}{frameworks}. 
Additionally, when combined with time series modeling techniques (e.g., Koopman operator, LSTM), the proposed technique offers significant improvements for surrogate modeling of high-dimensional multi-scale PDE systems. 
%\absbreak 

\end{abstract}
%%%%%%%%%%%%%%%%%%%%%%%%%%%
% \absbreak
%\rsbreak

%%%%%%%%%% Insert the texts which can accomdate on firstpage in the tag "fmtext" %%%%%

\section{Introduction}
Computational fluid dynamics involves solving large dynamical systems with millions of degrees of freedom, resulting in significant computational overhead. In order to alleviate this problem, reduced-order modeling \cite{deane1991low, LUCIA200451} is widely used, which uses a smaller number of modes to provide an approximate solution at a lower computational expense. A crucial step in reduced-order modeling is the projection of \textcolor{black}{high-dimensional} system states to a reduced latent space \cite{taira2017modal}. \textcolor{black}{The quality of projection can determine the overhead error for reduced-order modeling \cite{rowley2005model}}. Linear dimensionality reduction techniques such as Proper Orthogonal Decomposition (POD) \cite{sirovich1987turbulence,weller2010numerical,buoso2022stabilized, mezic2005spectral} are often used to create efficient representations of large-scale systems by projecting the solution manifold onto the space spanned by a set of linear orthonormal basis.

Advances in deep learning techniques, such as deep autoencoders (AE) \cite{hinton2006reducing}, capture intrinsic non-linear features for better compression, outperforming POD in low ranks, thus overcoming the well-known Kolmogorov barrier~\cite{peherstorfer2022breaking,ahmed2020breaking}. \textcolor{black}{Kolmogorov barrier is defined as a slow decay of the Kolmogorov \(n\)-width~\cite{cohen2015kolmogorovwidthsholomorphicmappings, MADAY2002289}, given by \( d_n(\mathcal{M}) = \inf_{\substack{U_n \subset \mathcal{V}, \dim(U_n) = n}} \sup_{x(\cdot; t, \mu) \in \mathcal{M}} \inf_{\hat{x} \in U_n} \| x(\cdot; t, \mu) - \hat{x} \| \), where \(\mathcal{M}\) is the solution manifold, \(\mathcal{V}\) is the ambient Hilbert space, \(U_n\) is any \(n\)-dimensional subspace of \(\mathcal{V}\), \(x(\cdot; t, \mu)\) is a solution to a parameterized PDE at time \(t\) and parameter \(\mu\), \(\hat{x}\) is its projection onto \(U_n\), and \(\|\cdot\|\) denotes the norm in \(\mathcal{V}\). The \(n\)-width quantifies the minimum worst-case projection error achievable by any \(n\)-dimensional linear subspace. This slow decay limits the best achievable error when using linear projection-based model reduction.} Milano and Koumoutsakos~\cite{milano2002neural} highlighted one of the first works to utilize a fully connected autoencoder to reconstruct the flow field, offering better performance compared to POD. Further studies have reported the usage of convolutional neural networks on 2D or 3D flow fields~\cite{pawar2019deep,zhang2023nonlinear,raj2023comparison}. Such methods have been adopted in the fluid dynamics community to obtain a nonlinear model order reduction~\cite{lee2020model, solera2024beta}, but they do not provide projection error convergence \textcolor{black}{as the rank of the latent space increases}. Recently, there have also been studies on using the hybrid approach \cite{dar2023artificial,barnett2023neural} combining POD with deep learning, by passing the latent space produced by POD to a neural network to find the corrections required to enhance reconstruction. 
These hybrid techniques have proven to enhance reconstruction beyond the capabilities of vanilla autoencoders. \textcolor{black}{Unlike these approaches, which treat POD as a fixed preprocessing step and use neural networks only to adjust the POD output, in this work} we propose a novel dimensionality reduction technique that combines traditional dimensionality reduction technique, i.e., POD, with deep learning techniques in a weighted manner at the encoder and decoder stage, where such weights of hybridization are also learned from data, to achieve a more effective dimensionality reduction. We also compare this approach with a straightforward hybrid technique, where a direct sum of POD and AE is utilized to construct the autoencoder, demonstrating the need of using learnable weighting parameters between POD and AE. 
Interestingly, our proposed approach obtains flat minima as opposed to other approaches, which contributes to the improved generalization and noise robustness. 

Building on our proposed framework, we also demonstrate the application of our framework to surrogate modeling tasks, particularly in the context of PDE system evolution.  We explore two distinct applications of our framework. First, we integrate our framework with the Koopman operator~\cite{mezic2021koopman,mezic2005spectral,arbabi2017ergodic}, where our framework facilitates the mapping of Koopman embedding back to physical space, while learning a linear forward model to evolve the system states in a reduced space. Second, we leverage long-short-term memory (LSTM) for \textcolor{black}{latent state} evolution \cite{hochreiter1997long}. LSTMs are usually trained on \textcolor{black}{the trajectories of} reduced states, obtained via POD \cite{wang2020recurrent, mohan2019compressed} or a non-linear autoencoder \cite{cae_lstm_rom_romit, gonzalez2018deep} to serve as a surrogate for the temporal evolution of latent states. The computational costs can be prohibitive when using LSTMs on full states, which can typically be in the range of millions for complex scenarios in fluid dynamics. By integrating our improved dimensionality reduction technique with a forward model for the \textcolor{black}{latent states}, we show that the \textcolor{black}{overall framework}  leads to a more accurate prediction of the system dynamics and provide insights into the individual error contributions from the dimensionality reduction and time series model. This underscores the fact that the quality of the reduced representation is the primary bottleneck in achieving highly accurate predictions, as suboptimal dimensionality reduction inherently limit the effectiveness of any downstream models. 

The remainder of the manuscript is structured as follows. In Section \ref{sec:method} we present the details of our hybrid dimensionality reduction technique, Koopman operator used for long term forecast, and surrogate modeling for learning PDE dynamics with LSTM. Section \ref{sec:Dataset_ae} describes the datasets used in each of the tasks, network architecture, and training hyper-parameters. Section \ref{sec:disc} provides a comprehensive evaluation of the proposed framework with respect to pure dimensionality reduction tasks and other downstream applications, comparing its performance against other techniques, and examining some key properties related to sharpness and noise robustness. 
Finally, Section \ref{sec::conc} concludes the paper. 

\section{Methodology}\label{sec:method}
\subsection{Dimensionality Reduction}
Without loss of generality, we begin by sampling a general vector-valued spatial-temporal field $u(x,t) \in \mathbb{R}^{Q}$ on a fixed mesh with $N$ cells, where $(x,t)$ is the space-time coordinate. At each time $t$, a cell-centered snapshot sample is a matrix $\mathbf{x} \in \mathbb{R}^{N \times Q}$.  
In the current framework, we start with two encoders: 
\begin{enumerate}
\item 
POD based encoder using $r$-dominant left singular vectors from the matrix consisting of stacked flattened columns of $\mathbf{x}$, denoted as $\phi_{\textrm{POD}}$. 
% as shown in \Cref{eq:pod}; 
% \begin{equation}
% \label{eq:pod}
%     \begin{bmatrix}
%     \mathbf{x}_1 & 
%     \mathbf{x}_2 & 
%     \ldots & 
%     \mathbf{x}_M
%     \end{bmatrix} \approx 
%     \mathbf{U}_{r} \Sigma_{r} \mathbf{V}_r^\top,
% \end{equation}
% where $\{\mathbf{x}_i\}_{i=1}^{M}$ is the set of $M$ training data, e.g., collection of flowfield snapshots;  
\item 
the neural network encoder with output dimension as $r$, denoted as $\phi_{\textrm{NN}}$. 
As shown in \Cref{eq:hybrid_encode}, the latent state $\mathbf{z} \in \mathbb{R}^{r}$ is obtained by a weighted sum of POD projection and the output of encoder, 
\begin{equation}
\label{eq:hybrid_encode}
    \textcolor{black}{\mathbf{z} = (\mathbf{1}-\mathbf{a})\odot\phi_{\textrm{POD}}(\mathbf{x}) + \mathbf{a}\odot\phi_{\textrm{NN}}(\mathbf{x}),}
\end{equation}
where $\mathbf{1}\in\mathbb{R}^{r}$ is a vector of ones, and $\mathbf{a}\in\mathbb{R}^r$ is a vector of learnable weights. \textcolor{black}{$\phi$ is a function that produces a vector in $r$ dimensional subspace. The weight $\mathbf{a}$ is multiplied via a element-wise multiplication ($\odot$) to the latent representation produced by POD and NN.} 
\end{enumerate}
Next, for the decoder part, we project the latent state $\mathbf{z}$ back to the reconstructed system state following \Cref{eq:hybrid_decode},  
\begin{equation}
\label{eq:hybrid_decode}
    \textcolor{black}{\hat{\mathbf{x}} = \psi_{\textrm{POD}}(\mathbf{z})\odot(\mathbf{1}-\mathbf{b}) + \psi_{\textrm{NN}}(\mathbf{z})\odot\mathbf{b},}
\end{equation}
\textcolor{black}{where $\mathbf{1}\in\mathbb{R}^{Q}$ is a vector of ones,  $\mathbf{b}\in\mathbb{R}^{Q}$, $\psi_{\textrm{POD}}$, and $\psi_{\textrm{NN}}$ are POD decoder and NN decoder, respectively.} \textcolor{black}{$\psi$ projects the latent vector into a physical space of dimension $\mathbb{R}^{N\times Q}$. Further an element wise multiplication ($\odot$) along the axis with dimension $Q$ multiplies the weights for the POD and NN part and combines them via an element wise addition}.
In addition to the parameters of NN encoder and decoder, both $\mathbf{a}\in\mathbb{R}^{r}$ and $\mathbf{b}\in\mathbb{R}^{Q}$ are trainable through gradient-based optimization as well. Hence, we name the above framework as \emph{learnable weighted hybrid autoencoder}. It is important to note that such NN can be either fully-connected or convolutional. 
% talk about loss function

Given the training dataset $\mathcal{D} = \{\mathbf{x}_i\}_{i=1}^{M}$, we trained the autoencoder by minimizing the mean-squared error (MSE) $\min_{\mathbf{\Theta},\mathbf{a},\mathbf{b}} \frac{1}{M} \sum_{i=1}^{M} \Vert \mathbf{x}_i - \hat{\mathbf{x}_i} \Vert^2$, 
where $\Vert \cdot \Vert$ is Frobenius norm, and $\mathbf{\Theta}$ refers to the set of the trainable parameters of neural network encoder $\phi_{\textrm{NN}}$ and decoder $\psi_{\textrm{NN}}$. 
$\mathbf{\Theta}$ is initialized using standard Kaiming~initialization \cite{he2015delving}. Motivated by Wang et al. \cite{Wang2024PirateNetsPD}, we choose to initialize $\mathbf{a}$ and $\mathbf{b}$ with zeros, leading to the proposed framework being equivalent to the classical POD at the beginning of neural network training. Thus, the model starts from the optimal linear encoder and becomes progressively nonlinear as the training proceeds. We choose Adam optimizer with learning rate of $10^{-4}$ for $\mathbf{\Theta}$ and $10^{-5}$ for $\mathbf{a}$ and $\mathbf{b}$.

\begin{figure}[H]
\centering
\includegraphics[width=\textwidth]{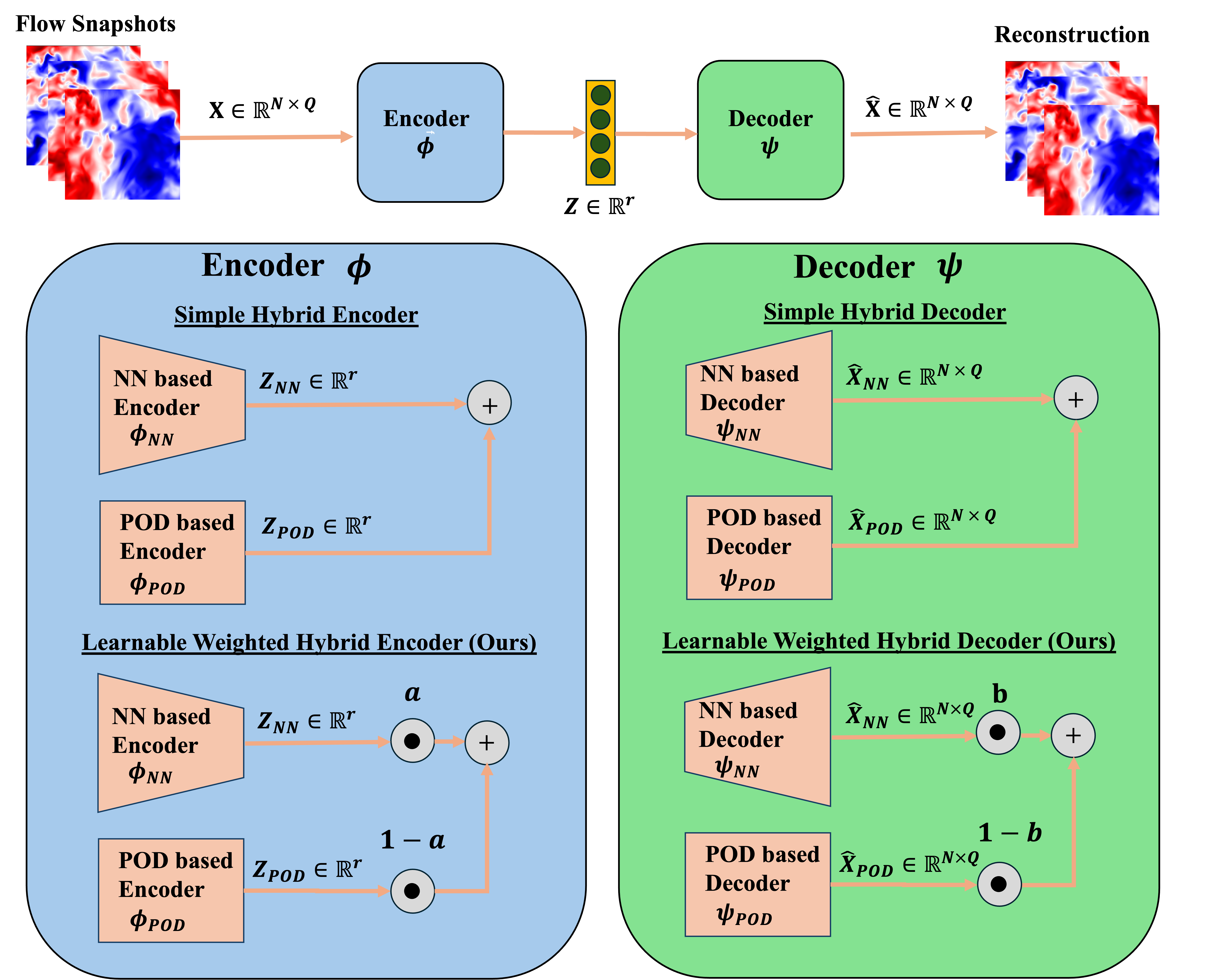}
%\rmfamily
\caption{\textcolor{black}{Architecture of the simple hybrid autoencoder and learnable weighted hybrid autoencoder.}}
\label{fig:top_figure}
\end{figure}

To emphasize the role of learnable weight $\mathbf{a}$ and \textcolor{black}{$\mathbf{b}$}, we also implement a straightforward hybrid approach~\cite{kosut2021quantum}, which simply adds the latent states from POD and NN. Similarly, the state of the reconstructed system is the sum of the output of the POD decoder and the NN decoder. We name this hybrid approach as \emph{simple hybrid autoencoder}. The difference of two approaches is further illustrated in \Cref{fig:top_figure}. 
As we shall see in the following sections, this comparison underscores the importance of using a weighted blend of the encoded and decoded spaces, as without it, the improvement over POD is at most incremental. 
% 
% The latent space and reconstruction of the simple hybrid approach is given by
% \begin{equation}{\label{hybrid_simple_encode}}
%     \mathbf{z}_{simple}(\boldsymbol{\beta}) = \boldsymbol{\phi}_{POD}(\mathbf{x}) + \boldsymbol{\hat{\phi}}_{NN}(\mathbf{x})
% \end{equation}
% \begin{equation}{\label{hybrid_simple_decode}}
%     \mathbf{\Tilde{x}}_{simple} = \boldsymbol{\psi}_{POD}(\mathbf{z}_{simple}) + \boldsymbol{\hat{\psi}}_{NN}(\mathbf{z}_{simple})
% \end{equation}
% In this case, only the weights and biases of the NN based encoder and decoder are the learnable parameters.
% 

% \begin{figure*}[ht]
% \centering
% \subcaptionbox{NN based encoder-decoder utilizing fully connected neural networks}{\includegraphics[width=0.3\textwidth]{figs/nn_based.png}}%
% \hfill % <-- Separation
% \hfill
% \subcaptionbox{NN based encoder-decoder utilizing convolutional neural networks}{\includegraphics[width=0.7\textwidth]{figs/CNN_arch.png}}%
% \caption{Different types of NN based encoder-decoder architecture}
% \label{archs}
% \end{figure*}

% here 

\subsection{Koopman Forecasting}
Koopman theory \cite{koopman1931hamiltonian} states that any nonlinear dynamical system can be linearized by lifting into a (possiblly infinite-dimensional) space of observable functions. Hence, given a continuous dynamical system $\dot{\mathbf{x}} = \mathbf{f}(\mathbf{x})$, for any observable $\xi$, one defines the Koopman operator $\mathcal{K}$ as:
\begin{equation}
\xi(\mathbf{x}(t+\Delta t)) =  \mathcal{K} \xi(\mathbf{x}(t)), 
\end{equation}
where $t$ refers to the time and $\Delta t$ is the time step size. Although the idea of linear evolution is attractive and desirable in most scenarios, the effectiveness of Koopman theory is often limited by the subspace chosen.

To overcome these challenges, Lange et al.~\cite{lange2021fourier} proposed \emph{Koopman forecasting}, 
\begin{equation}
\xi(\mathbf{x}(t)) =  
\begin{bmatrix} 
\cos(\vec{\mathbf{\omega}} t) \\ 
\sin(\vec{\mathbf{\omega}} t) 
\end{bmatrix} 
:= \Omega(\vec{\mathbf{\omega}} t),
\end{equation}
\textcolor{black}{where $\vec{\omega}$ denotes the frequency vector of dimension $N_f$, corresponding to the number of distinct frequencies considered. Consequently, the rank of the reduced representation, according to the formulation, becomes $r = 2N_f$}. With these assumptions the system state at any time $t$ is given by 
\begin{equation}
    \mathbf{x}(t) = \psi_{\theta}(\Omega(\vec{\omega} t)).
\end{equation}
We solve the following minimization problem to learn the non-linear mapping $\psi_{\theta}$ and frequencies $\vec{\omega}$, where $\psi_\theta$ is parametrized by $\theta$ which includes the neural network's weights, biases and the learnable weighting parameters :
\begin{equation}
\min_{\vec{\omega}, \theta} \sum_{i=1}^{T} \left\| \mathbf{x}_i - \psi_{\theta} (\Omega(\vec{\omega} i\Delta t)) \right\|^2,
\end{equation}
where $i$ indicates the timestep index and $\Delta t$ is the timestep size between consecutive snapshots. We utilize our proposed hybrid framework to learn the mapping from observable space to the physical state, while keeping their time evolution strategy unchanged, taking the same form as in \Cref{eq:hybrid_decode}
\begin{equation}
\label{eq:koopman_decode}
    \textcolor{black}{\psi_{\theta}(\Omega(\vec{\omega} t)) = \psi_{\textrm{POD}}(\Omega(\vec{\omega} t))\odot(\mathbf{1}-\mathbf{b}) + \psi_{\textrm{NN}}(\Omega(\vec{\omega} t))\odot\mathbf{b},}
\end{equation}
with $\mathbf{b}$ initialized to zeros during training. It should be noted that this problem involves only the learning of a decoder jointly with the Koopman model for time evolution.

\subsection{Surrogate Modeling for Time-Dependent PDEs}

Our surrogate modeling strategy involves the usage of the aforementioned techniques to obtain a latent representation and LSTM to determine its evolution over time. Training is divided into two stages. The dimensionality reduction framework is first trained to obtain a low-dimensional embedding of the system states, which subsequently serves as the training data for the time series prediction model to evolve the system in time\cite{cae_lstm_rom_romit, Halder_2024}. The advantage of training them separately is that it allows us to clearly demarcate the contribution of each stage to the overall surrogate model, which is particularly important as the crux of the current study is the introduction of a novel framework for dimensionality reduction. 

LSTM networks are a type of recurrent neural network (RNN) designed to address the vanishing gradient problem in standard RNNs. LSTMs incorporate a gating mechanism to selectively retain and propgate information over long sequences, making them suitable for modeling sequential and time-dependent data. They are auto-regressive in nature, using data from the past \( k \) timesteps to predict the next state. To predict \( x_{i+1} \), the model uses information from \( x_i, x_{i-1}, \ldots, x_{i-k+1} \), effectively capturing temporal dependencies in multiple time steps. $k$ is referred to as the look-back window and is usually tuned based on the problem and the dataset at hand. In the current study, we set the look-back window to a value of 10 for every dataset. This value was found to be optimal in our analysis. Other architectural details of the LSTM will be discussed on a case-by-case basis. 

\section{Datasets and model setup}\label{sec:Dataset_ae}
\subsection{Chaotic Fluid System}

% \subsection{Setup of Synthetic Data}
% In this work, we use two canonical chaotic PDEs: 1D Kuramoto-Sivashinsky and homogeneous isotropic turbulence datasets for numerical experiments. 
% \subsection{1D Viscous Burger's}
% The general form of 1D viscous Burger's equation is given by
% \[
% \frac{\partial u}{\partial t} + u \frac{\partial u}{\partial x} = \nu \frac{\partial^2 u}{\partial x^2}
% \]
% Random initial condition was used on a domain of size $2\pi$ to generate the data for around 3000 time steps using finite difference method. Grid sizes of 512, 1024 and 2048 was used for the simulation. The simulation data contains shock formation at different locations. \\
% The neural network architecture used for the NN based encoder and decoder for this dataset is a fully connected network with one hidden layer. Tanh activation was used at the input and hidden layer. The dataset was split into training and test set in a 70:30 manner. Scaling of the data was performed prior to training using the mean and standard deviation of the data. Adam optimizer was used with a learning rate of 1e-3 and scheduler to reduce the learning rate upon plateauing. 

\subsubsection{Kuramoto-Sivashinsky (KS)}
\label{sec:ks_description}

The KS equation is  given by$
\frac{\partial u}{\partial t} + u \frac{\partial u}{\partial x} + \frac{\partial^2 u}{\partial x^2} + \frac{\partial^4 u}{\partial x^4} = 0$, 
where $x \in [0, L_x)$, $L_x=64\pi$, and periodic boundary conditions are assumed. The initial conditions comprise the sum of ten random sine and cosine waves $u(x,0) = \sum_{k=1}^{10} A_k \left( \sin\left(\frac{2\pi n_k x}{L_x} + \phi_k\right) +\cos\left(\frac{2\pi n_k x}{L_x} + \phi_k\right)\right)$, where $\forall k \in \{1,\ldots,10\}, A_k \sim U(-1, 1),\phi_k \sim U(0, 2\pi)$ and wavenumber $n_k \sim U\{1,\ldots,6\}$, respectively.
We employ Fourier decomposition in space and \textcolor{black}{2nd order Crank-Nicolson/Adam-Bashforth} semi-implicit finite difference scheme for temporal discretization. The data is generated for around 2000 time steps with a timestep of 0.01. Grid sizes of 512, 1024 and 2048 are used for the simulation. Some initial snapshots of the simulation that correspond to the transient phase are ignored.

For the NN encoder, we use a feedforward neural network with one hidden layer with $2r$ number of neurons in the hidden layer. Hyperbolic tangent function is used for activation in all layers except the last linear layer. The NN decoder is symmetric to the encoder. After shuffling, the snapshot data is split into training and testing with a 7:3 ratio. Before training, we standardize the data by subtracting the mean and dividing by the standard deviation. All models are trained for 40k epochs with a batch size of 64.

\subsubsection{Homogeneous isotropic turbulence (HIT)}
\label{sec:hit}
Direct numerical simulation data of forced homogeneous isotropic turbulence is obtained from the Johns Hopkins turbulence database~\cite{li2008public}. The data set is generated by solving the forced Navier-Stokes equation on a periodic cubic box using the pseudospectral method. We interpolate the velocity field, $u(x,t)=(u_x,u_y,u_z)\in\mathbb{R}^3$,  from the original dataset of resolution $1024^3$ to grid sizes of $16^3$, $32^3$ and $64^3$ and extract the data from timestep 1 to 2048 with a stride of 16, resulting around 128 snapshots. 

% \begin{align}
% &\frac{\partial \mathbf{u}}{\partial t} + (\mathbf{u} \cdot \nabla) \mathbf{u} = -\nabla p + \nu \nabla^2 \mathbf{u} + \mathbf{f}, 
% &\nabla \cdot \mathbf{u} = 0,
% \end{align}

We choose a deep convolutional autoencoder~\cite{fukami2020convolutional} as the NN part of the proposed framework. The details of the architecture pertaining to the NN based encoder and decoder and the training hyper parameters are shown in \Cref{tab:nn_architecture}. Without shuffling, we systematically sample the dataset by selecting every alternate snapshot to curate the training and testing data, each comprising 64 snapshots.

\begin{table}[H]
    \centering
    \scriptsize
    \caption{Architecture details and training parameters for 3D HIT dataset}
    \label{tab:nn_architecture}
    \renewcommand{\arraystretch}{1.2} % Adjust row height for better readability
    \begin{tabular}{p{3cm} p{3cm} p{3cm} p{3cm}}
        \toprule
        \multicolumn{4}{c}{\textbf{Neural Network Architecture}} \\ 
        \midrule
        \multicolumn{2}{c}{\textbf{Encoder Architecture}} & \multicolumn{2}{c}{\textbf{Decoder Architecture}} \\
        \midrule
        \textbf{Component} & \textbf{Details} & \textbf{Component} & \textbf{Details} \\
        \midrule
        Hidden Layers      & 4 Convolutional Layers     & Hidden Layers      & 4 Transpose Convolutional Layers   \\ 
        Filters            & 256, 512, 1024, 2048       & Filters            & 2048, 1024, 512, 256               \\ 
        Bottleneck Layer   & Fully Connected Linear     & Output Layer       & Fully Connected Linear             \\ 
        Activation Function& Swish (except bottleneck)  & Activation Function& Swish (except output)              \\ 
        Dropout            & 0.4 (all except bottleneck)& Dropout            & 0.4 (all except output)            \\ 
        \midrule
        \multicolumn{4}{c}{\textbf{Training Hyperparameters}} \\
        \midrule
        \multicolumn{2}{l}{\textbf{Parameter}} & \multicolumn{2}{l}{\textbf{Value}} \\
        \midrule
        \multicolumn{2}{l}{Epochs}     & \multicolumn{2}{l}{2000} \\
        \multicolumn{2}{l}{Batch Size} & \multicolumn{2}{l}{20}   \\
        \multicolumn{2}{l}{Optimizer}  & \multicolumn{2}{l}{Adam} \\
        \multicolumn{2}{l}{Learning Rate} & \multicolumn{2}{l}{$1\times10^{-4}$} \\
        \bottomrule
    \end{tabular}
\end{table}

% The loss function being minimized here is similar to \Cref{minimization}, but it needs to include the 3 dimensions and channels of the dataset, described as
% \begin{align}
% \nonumber
% \mathcal{L}_{\text{MSE}}(\theta) &= \frac{1}{3 N N_x N_y N_z} \sum_{i=1}^{N}
% \sum_{j_1=1}^{N_x} \sum_{j_2=1}^{N_y}
% \sum_{j_3=1}^{N_z}
% \bigg( \\
% \nonumber
% & (u_{i,j_1,j_2,j_3} - \tilde{u}_{i,j_1,j_2,j_3})^2 + (v_{i,j_1,j_2,j_3} - \tilde{v}_{i,j_1,j_2,j_3})^2 \\
% & + (w_{i,j_1,j_2,j_3} - \tilde{w}_{i,j_1,j_2,j_3})^2  \bigg)
% % \mathbf{x}^{(i,k,j_1,j_2,j_3)} - \\
% % & \mathbf{\tilde{x}}^{(i,k,j_1,j_2,j_3)})^2,
% \label{eq:mse_loss_3d_multiline}
% \end{align}
% where $N_x,N_y,N_z$ refers to the number of grid points in each direction of $x,y,z$, respectively.

\subsection{Koopman Forecasting Datasets}
\subsubsection{Traveling Wave}
We demonstrate our framework on a traveling wave problem with a spatial dimensionality of 256. The spatio-temporal evolution of the wave is given by the following: 
\begin{equation}
u(x,t) = \mathcal{N} \left( x \mid \mu = 100(\sin(0.01t) + 1) + 28, \sigma^2 = 10 \right), 
\end{equation}
where $\mathcal{N}$ is the probability density function of Gaussian distribution. 
We generate a trajectory of 100,000 timesteps, of which the initial 50,000 snapshots are used for training the Koopman-Decoder model, and the remaining 50,000 timesteps are held out for testing. The details of the non-linear part of the decoder and training parameters are provided in \Cref{tab:decoder_architecture_koopman_wave}. The Koopman model is parametrized by a single frequency, essentially evolving the dynamics in a latent space of rank 2.

\begin{table}[h]
    \centering
    \scriptsize % Makes the text smaller for compact formatting
    \caption{Decoder Architecture and Training Hyperparameters for the traveling wave problem}
    \label{tab:decoder_architecture_koopman_wave}
    \renewcommand{\arraystretch}{1.2} % Adjust row height for readability
    \begin{tabular}{p{6cm} p{6cm}}
        \toprule
        \multicolumn{2}{c}{\textbf{Decoder Architecture}} \\
        \midrule
        \textbf{Component} & \textbf{Details} \\
        \midrule
        Initial Layer & Fully Connected Linear Layer \\
        Hidden Layers & 4 Transpose Convolutional Layers \\
        Channels & 32, 16, 8, 4 \\
        Activation Function & SiLU (except after linear and final layer) \\
        \midrule
        \multicolumn{2}{c}{\textbf{Training Hyperparameters}} \\
        \midrule
        \textbf{Parameter} & \textbf{Value} \\
        \midrule
        Optimizer & Adam \\
        Learning Rate & \(3 \times 10^{-4}\) \\
        Batch Size & 1280 \\
        Epochs & 1000 \\
        \bottomrule
    \end{tabular}
\end{table}

\subsubsection{Flow Over Cylinder}
We utilize the time evolution data of vorticity in a two-dimensional flow over a cylinder at $Re\approx100$~\cite{lange2021fourier}. The initial 50 temporal snapshots are used in training the Koopman-Decoder model, and the remaining 100 are kept out for testing. The details of the non-linear part of the decoder and training parameters are provided in \Cref{tab:decoder_architecture_koopman_cylinder}. The Koopman model is parametrized by two frequencies, essentially evolving the dynamics in a latent space of rank 4.

\begin{table}[h]
    \centering
    \scriptsize % Makes the text smaller for compact formatting
    \caption{Decoder Architecture and Training Hyperparameters for the flow over cylinder problem}
    \label{tab:decoder_architecture_koopman_cylinder}
    \renewcommand{\arraystretch}{1.2} % Adjust row height for readability
    \begin{tabular}{p{6cm} p{6cm}}
        \toprule
        \multicolumn{2}{c}{\textbf{Decoder Architecture}} \\
        \midrule
        \textbf{Component} & \textbf{Details} \\
        \midrule
        Initial Layer & Fully Connected Linear Layer \\
        Hidden Layers & 4 Transpose Convolutional Layers \\
        Channels & 256, 128, 64, 32 \\
        Activation Function & SiLU (except after linear and final layer) \\
        \midrule
        \multicolumn{2}{c}{\textbf{Training Hyperparameters}} \\
        \midrule
        \textbf{Parameter} & \textbf{Value} \\
        \midrule
        Optimizer & Adam \\
        Learning Rate & \(3 \times 10^{-4}\) \\
        Batch Size & 8 \\
        Epochs & 1000 \\
        \bottomrule
    \end{tabular}
\end{table}

\subsection{Surrogate Modeling Datasets}
\subsubsection{1D Viscous Burgers' Equation}
The one-dimensional Viscous Burgers' equation is represented by the following partial differential equation.
\begin{equation}
\frac{\partial u}{\partial t} + u \frac{\partial u}{\partial x} = \nu \frac{\partial^2 u}{\partial x^2},
\end{equation}

\begin{equation}
u(x,0) = u_0, \quad x \in [0, L], \quad u(0,t) = u(L,t) = 0.
\end{equation}

The initial condition used is given by $u(x,0) = x/\left(1 + \sqrt{\frac{1}{t_0}} \exp \left( Re\frac{x^2}{4} \right)\right)$. This initial condition allows for an analytical solution of the form $u(x,t) = \frac{x}{t+1}/\left({1 + \sqrt{\frac{t+1}{t_0}} \exp \left( Re \frac{x^2}{4(t+4)} \right)}\right)$, where $t_0 = e^{Re/8}$ and (Reynolds Number) $Re=\frac{1}{\nu}$. Data generation is carried out by computing the analytical solution on a grid of 128 points for 100 timesteps with a terminal time of 2s. Model training utilizes trajectories of 19 different $Re$ values ranging from 100 to 1900 in steps of 100, with 100 time steps in each. For testing, we hold out trajectories of 13 different $Re$ values ranging from 50 to 2450 in steps of 200. 

A fully connected network using the same architecture as outlined in \Cref{sec:ks_description} is used for the NN based encoder and decoder. The dimensionality reduction framework was trained with a learning rate of $1\times 10^{-3}$, batch size of 32 for 500 epochs using the Adam optimizer. To enhance convergence, we used a cyclic learning rate scheduler, which dynamically adjusts the learning rate during training. A latent space dimension of 2 is chosen for this problem.

Following the training and convergence of the dimensionality reduction framework, the encoder is utilized to obtain the reduced-order representations. The resulting embeddings, 
denoted as $\mathbf{z} \in \mathbb{R}^r$, are subsequently augmented with the $Re$ values, thereby extending the latent space dimension from $r = 2$ to $r' = 3$. This enriched representation, 
$\tilde{\mathbf{z}} = [\mathbf{z}, Re]$, serves as the input to the LSTM model, where the objective is to approximate the discrete-time mapping $\tilde{\mathbf{z}}_{i+1} = \mathcal{F}_{\theta}(\tilde{\mathbf{z}}_{i}, Re)$, parameterized by $\theta$ (LSTM weights and biases of the NN). At this stage, the emphasis shifts from state compression to modeling the underlying temporal dynamics in the reduced-order manifold. The LSTM is then trained to learn the evolution of $\tilde{\mathbf{z}}$, facilitating the 
construction of a surrogate model capable of forecasting trajectory evolution given initial conditions and governing parameters.

The LSTM consists of 2 hidden layers with 40 neurons~\cite{cae_lstm_rom_romit}. The model is trained with a batch size of 32 for 400 epochs with a learning rate of $10^{-3}$ using Adam optimizer along with a cyclic learning rate scheduler. Data scaling is not performed at any stage of the training. The task here is to learn a parametric surrogate model capable of capturing the system state evolution via a reduced representation.

% \begin{table}[H]
%     \centering
%     \scriptsize % Makes text compact for better readability
%     \caption{Neural Network Architecture and Training Hyperparameters for 1D Viscous Burger's case}
%     \label{tab:1d_vb_arch}
%     \renewcommand{\arraystretch}{1.2} % Adjust row height for readability
%     \begin{tabular}{p{3cm} p{3cm} p{3cm} p{3cm}}
%         \toprule
%         \multicolumn{4}{c}{\textbf{Neural Network Components}} \\
%         \midrule
%         \multicolumn{2}{c}{\textbf{Encoder-Decoder}} & \multicolumn{2}{c}{\textbf{LSTM}} \\
%         \midrule
%         \textbf{Component} & \textbf{Details} & \textbf{Component} & \textbf{Details} \\
%         \midrule
%         Hidden layers & Same as used in \Cref{sec:ks}  & Hidden Layers  & 2 with 40 neurons per layer \\
%         Latent Space Dimension & 2 & Input Dimension & 3 \\
%         Activation & SiLU (except final layer) &  &  \\
%         \midrule
%         \multicolumn{4}{c}{\textbf{Training Hyperparameters}} \\
%         \midrule
%         \textbf{Parameter} & \textbf{Value} & \textbf{Parameter} & \textbf{Value} \\
%         \midrule
%         Learning Rate & \(1 \times 10^{-3}\) & Learning Rate & \(1 \times 10^{-3}\) \\
%         Batch Size & 32 & Batch Size & 32 \\
%         Epochs & 500 & Epochs & 400 \\
%         Optimizer & Adam & Optimizer & Adam \\
%         Scheduler & Cyclic Learning Rate & Scheduler & Cyclic Learning Rate \\
%         \bottomrule
%     \end{tabular}
% \end{table}

\subsubsection{2D Shallow Water Equations}
We extend the surrogate modeling to two dimensions by using inviscid shallow water equations of the form
\begin{align}
    \frac{\partial (\rho \eta)}{\partial t} + \frac{\partial (\rho u \eta)}{\partial x} + \frac{\partial (\rho v \eta)}{\partial y} &= 0, \\[10pt]
    \frac{\partial (\rho u)}{\partial t} + \frac{\partial}{\partial x} \left( \rho u^2 + \frac{1}{2} \rho g \eta^2 \right) + \frac{\partial (\rho u v)}{\partial y} &= 0, \\[10pt]
    \frac{\partial (\rho v)}{\partial t} + \frac{\partial (\rho u v)}{\partial x} + \frac{\partial}{\partial y} \left( \rho v^2 + \frac{1}{2} \rho g \eta^2 \right) &= 0,
\end{align}
with $\eta$ representing the fluid column height, ($u,v$) referring to the depth averaged horizontal and vertical velocity of the fluid, $\rho$ being density of the fluid and $g$ is the acceleration due to gravity. The fluid density is kept constant at a value of 1.0. The governing equations are solved in a square domain of unit dimension, discretized using 64 grid points in each direction, periodic boundary conditions are enforced and initial conditions being
\begin{align}
    \rho \eta (x,y,t=0) &= \exp \left( - \frac{(x-x_0)^2}{0.005} - \frac{(y-y_0)^2}{0.005} \right), \\
    \rho u (x,y,t=0) &= 0, \\
    \rho v (x,y,t=0) &= 0.
\end{align}
The initial conditions refer to a Gaussian pulse in the fluid column height, where the parameters $x_0$ and $y_0$ denote the initial location of the Gaussian pulse. We generate 90 trajectories for training, each representing a different initial location of pulse with a timestep of 0.001 s. The simulation is run for a final time of 0.5 s with the states being saved every 5 timesteps, leaving us 100 timesteps in each trajectory. Ten other trajectories are held for testing purposes. 

A deep convolutional autoencoder is used for the NN part of the framework with architecture as outlined in \Cref{sec:hit}, with the only difference being that convolution and deconvolution operations were performed on two-dimensional data. The dimensionality reduction framework was trained first with a learning rate of $3\times 10^{-4}$, batch size of 24 for 500 epochs using Adam optimizer along with cyclic learning rate scheduler. We set the latent space rank to $r=6$. The full order states are scaled to zero mean and unit variance.
Similar to the 1D Viscous Burgers case, we augment the latent space variable before training the LSTM model. Since this is a nonparametric PDE, the latent space variable here is augmented with the initial location of the pulse $\bar{x}, \bar{y}$. This takes the LSTM input dimension from 6 to 8 for every trajectory. The LSTM consists of 3 hidden layers with 50 neurons. The model is trained with a batch size of 24 for 400 epochs with a learning rate of $10^{-3}$ using Adam optimizer along with a cyclic learning rate scheduler. The task here is to learn a surrogate model in reduced space given the initial condition and location of the pulse.

\subsubsection{3D Viscous Burgers' Equations} 
A straightforward extension of the 1D Viscous Burgers' into three dimensions gives us the three-dimensional Viscous Burgers' equation of the form

\begin{equation}
\frac{\partial \mathbf{u}}{\partial t} = - b \frac{1}{2} \nabla \cdot (\mathbf{u} \otimes \mathbf{u}) + \nu \nabla \cdot \nabla \mathbf{u},
\end{equation}
where $\mathbf{u} = (u, v, w)$ is the velocity field, $b$ is the advection parameter and $\nu$ is the diffusion parameter. 
Simulation data for the 3D Viscous Burgers' equation is obtained from APEBench dataset \cite{koehler2024apebench}. The advection and diffusion parameters are set to a value of -1.5 and 1.5 respectively, emulating a decaying dynamics. We generate the simulation data for 50 different trajectories, each with a different initial condition on a grid of size $32^3$ containing 101 temporal snapshots in each trajectory. For this dataset, we use the first 50 timesteps for training and the rest for testing purposes.

We use a deep convolutional autoencoder for the NN part of the framework with the same architecture as outlined in \Cref{sec:hit}. The dimensionality reduction framework was trained with a learning rate of $1\times 10^{-4}$, batch size of 20 for 3000 epochs using Adam optimizer. We choose a six-dimensional latent space to represent the system. The full orders states are scaled to zero mean and unit variance prior to training. 

For this problem, we do not augment the latent space with any control parameters prior to training the LSTM model. The LSTM consists of 3 hidden layers with 128 neurons. The learning hyperparameters are similar to the 2D Shallow Water case. In this case we evaluate the model's ability to extrapolate in time using the latent representation, which can be challenging given the dynamics are decaying.

\section{Results and Discussions}\label{sec:disc}

\subsection{Dimensionality Reduction Performance}

\begin{figure}[htbp]
\centering
%\hspace*{-0.20cm}
\includegraphics[width=\textwidth]{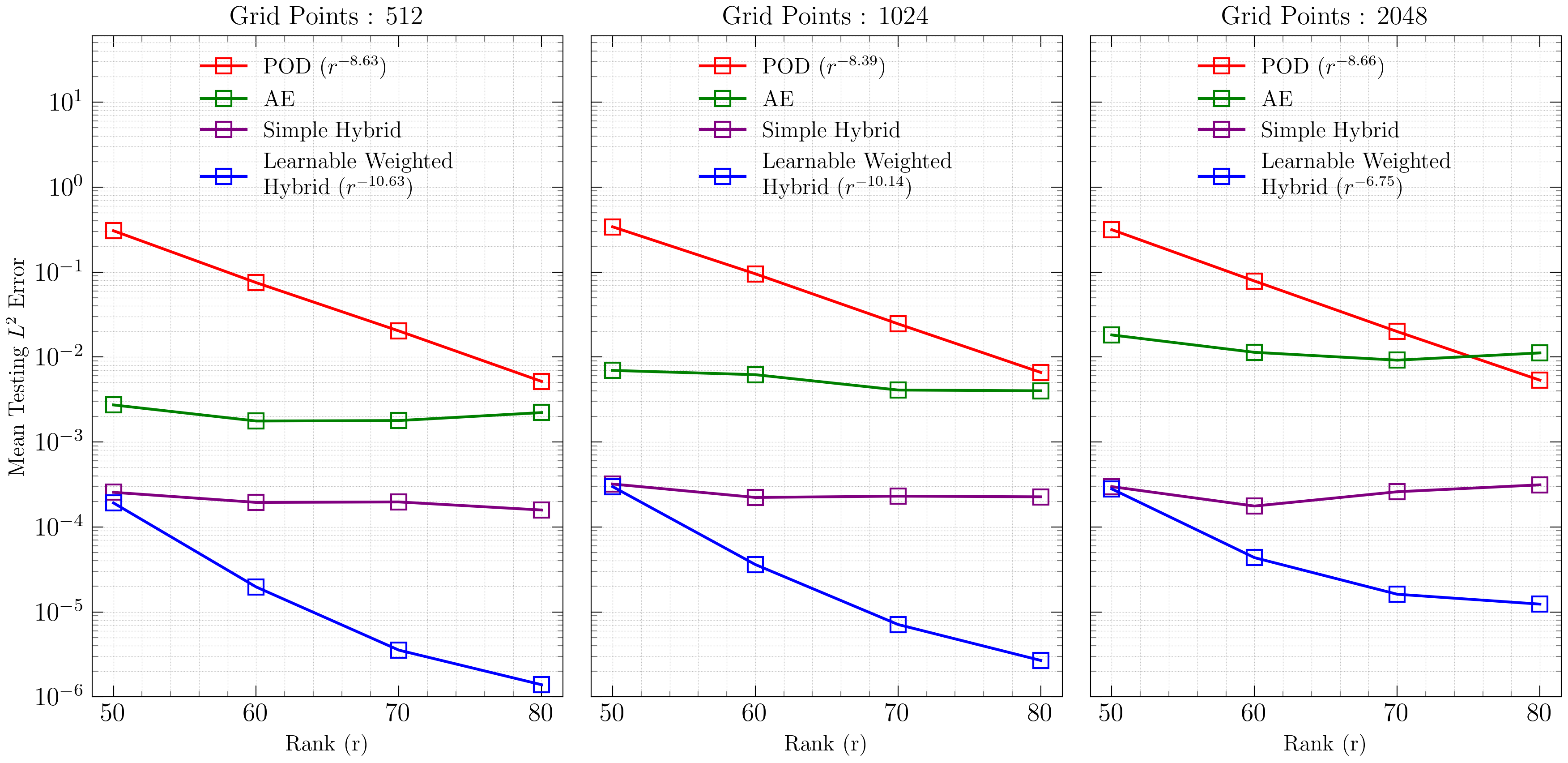}
%\rmfamily
\caption{Generalization performance of four models on 1D Kuramoto-Shivaskinsky dataset with varying rank $r$ and resolution $N$. X axis denotes the rank of the latent space and Y axis denotes the Mean Testing $L^2$ error. For each grid, rank, and method, the model is trained across 16 independent runs with varying random seed values. The results from these runs are averaged to compute the mean testing $L^2$ error. The convergence rate for POD and Learnable weighted hybrid approach is indicated in the legend as an exponent to $r$.}
\label{fig:1d_ks}
\end{figure}

To demonstrate the effectiveness of the proposed approach, we compare our model against three other methods. POD, vanilla deep autoencoder (AE), simple hybrid autoencoder, as autoencoders for the two chaotic PDE datasets in \Cref{sec:ks_description,sec:hit} with varying rank $r$ and resolution $N$. Sixteen ensembles of the model are trained by setting different random seeds. From \Cref{fig:1d_ks}, both of the two hybrid approaches perform better than AE and POD. However, \emph{simple hybrid approach} does not maintain its convergence\footnote{\textcolor{black}{We use the term \emph{convergence} in an empirical, asymptotic sense—referring to the monotonic decrease of training and generalization errors as the latent dimension $r$ increases. For sufficiently large $r$, the error tends toward a small bound $\epsilon$, exhibiting a qualitative decay resembling $\mathcal{O}(r^{-q})$, for some $q\in\mathbb{R}^{+}$. This is commonly desirable in reduced-order modeling, though perfect monotonicity may not hold—especially for regular autoencoders lacking optimal spectral alignment. Our learnable weighted model exhibits this trend and is bounded from below by POD, which often converges faster due to its optimality. Deviations may arise due to limited observability in coarse-grained turbulent fields.}} with increasing rank in contrast to our proposed approach. In addition, our approach shows an order-of-magnitude improvement in generalization as compared to any of the other methods. The simple and our learnable weighted hybrid approach give nearly the same test error at low ranks, but their gap increases multiple times with increasing rank. Surprisingly, such substantial improvement merely requires a negligible additional trainable parameters (i.e. $r+Q \ll N$), as shown in \Cref{tab:ks_table}. For this dataset the convergence of our approach follows a SVD-like convergence, while the simple approach has a behavior similar to AE.

% 
% For the K-S dataset from \Cref{sec:ks}, 
% \Cref{tab:ks_table} summarizes the mean training error, mean test error, and associated standard deviation, and the total number of model parameters. We report the error here as the $L^2$ distance between the ground truth field $u(\cdot,t)$ and reconstructed field. The results from the current approach are displayed in bold. 

% \Cref{fig:1d_ks} shows the test error with respect to rank for the same. Both the hybrid approaches perform better than AE and POD. But the simple approach doesn't maintain its convergence with rank, unlike the tunable approach, which also produces orders of magnitude improvement in test error as compared to any of the other methods. The simple and tunable (ours) hybrid approach gives nearly the same test error at low ranks, but their gap increases multi folds with increasing rank. The substantial improvement in error requires only a small number of additional trainable parameters, equivalent to the rank of the latent space. For this dataset the convergence of the tunable (ours) approach follows a SVD like convergence, while the simple approach has a behaviour similar to AE.

% 3D HIT discussion

\begin{figure}[htbp]
\centering
%\hspace*{-0.20cm}
\includegraphics[width=\textwidth]{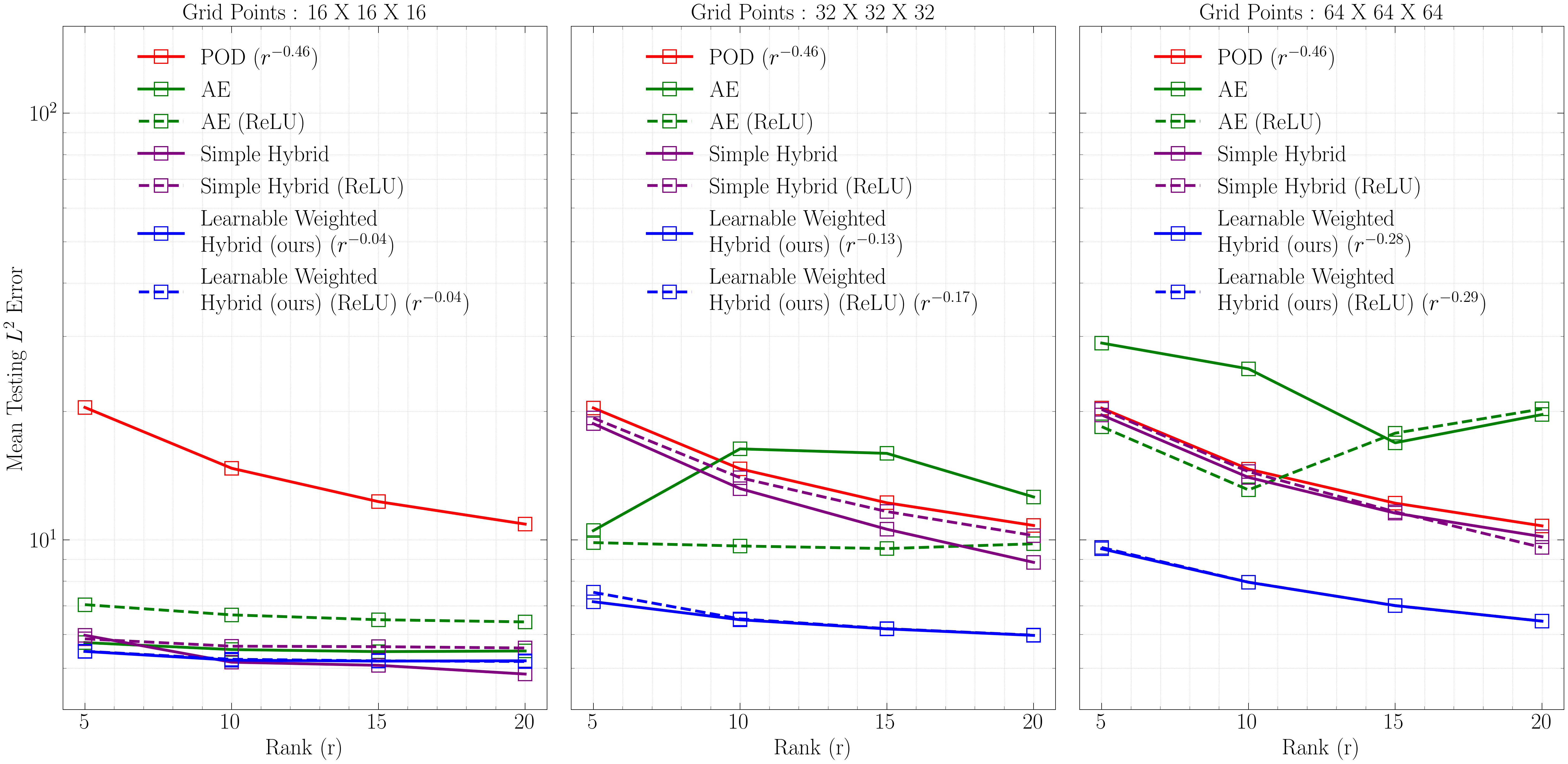}
%\rmfamily
\caption{Generalization performance of four models on 3D homogeneous isotropic turbulence dataset with varying rank $r$ and resolution $N$. The testing $L^2$ error obtained using ReLU activation function is indicated using dashed lines. Similar to K-S case, 16 independent runs with varying random seed values are performed to obtain the mean testing $L^2$ error.}
\label{fig:3d_hit}
\end{figure}

For the more challenging 3D HIT dataset, \Cref{fig:3d_hit} shows that our proposed approach continues to outperform the other three methods in terms of generalization, especially when the resolution increases (e.g., $32^3, 64^3$ as opposed to $16^3$). It is important to note that the simple approach shows little improvement over POD at resolutions of $32^3$ or $64^3$ while our approach excels. Again, this highlights the important role of learnable weights in our hybrid approach. Additionally, for this dataset we study the impact of activation function on the reconstruction performance by utilizing ReLU activation function for the NN part of the auto-encoder. It is worth noting that the performance of both the AE and simple hybrid approaches exhibit noticeable variations when the activation function is changed. In contrast, our approach shows minimal changes in performance, demonstrating the robustness of the proposed framework. \textcolor{black}{The latent space obtained using the proposed approach is also stable and reproducible (\Cref{sec:latentrep_robust}).}

\textcolor{black}{To highlight the state-of-the-art performance of the proposed framework, we also perform a one-off comparison of its generalization ability on the complex 3D HIT case against the $\beta$-VAE baseline \cite{solera2024beta}, as detailed in \Cref{sec:beta_vae}. Our framework consistently achieves superior reconstruction performance across all grid resolutions and latent space ranks.} \textcolor{black}{Note that the improvement in performance is obtained at little to no additional computational overhead as shown in \Cref{tab:hit_table} in terms of the total number of training parameters and the training time per epoch which be seen in \Cref{fig:wall_time} provided in \Cref{sec:train_time}.} 

\textcolor{black}{To better understand the relative roles of the linear POD basis and the nonlinear neural network in the proposed framework, we analyze their respective contributions to both the latent representation and the final reconstruction. A detailed breakdown of these contributions, across different resolutions and latent dimensions, is presented in \Cref{sec:rel_contrib}.} 

\begin{figure}[htbp]
\centering
%\hspace*{-0.1cm}
\includegraphics[width=\textwidth]{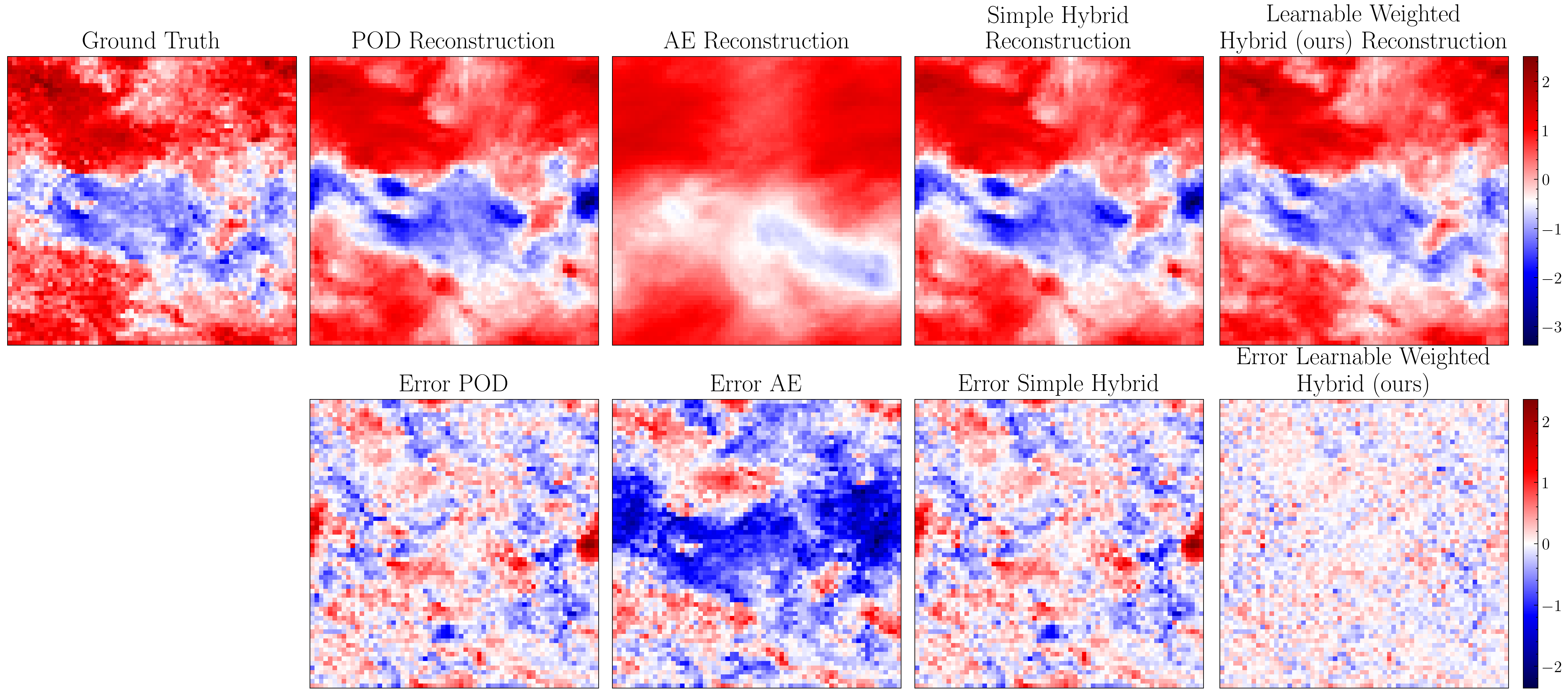}
%\rmfamily
\caption{Generalization performance comparison on 3D HIT data at a resolution of $64^3$ and rank $r=5$, with a 2D slice of $u_y$. Top: $u_y$. Bottom: absolute error of $u_y$ between the reconstructed field and ground truth.}
\label{fig:recon}
\end{figure}

\subsection{Koopman Forecasting Performance}
\begin{figure}[h!]
\centering
%\hspace*{-0.20cm}
\includegraphics[width=1.0\textwidth]{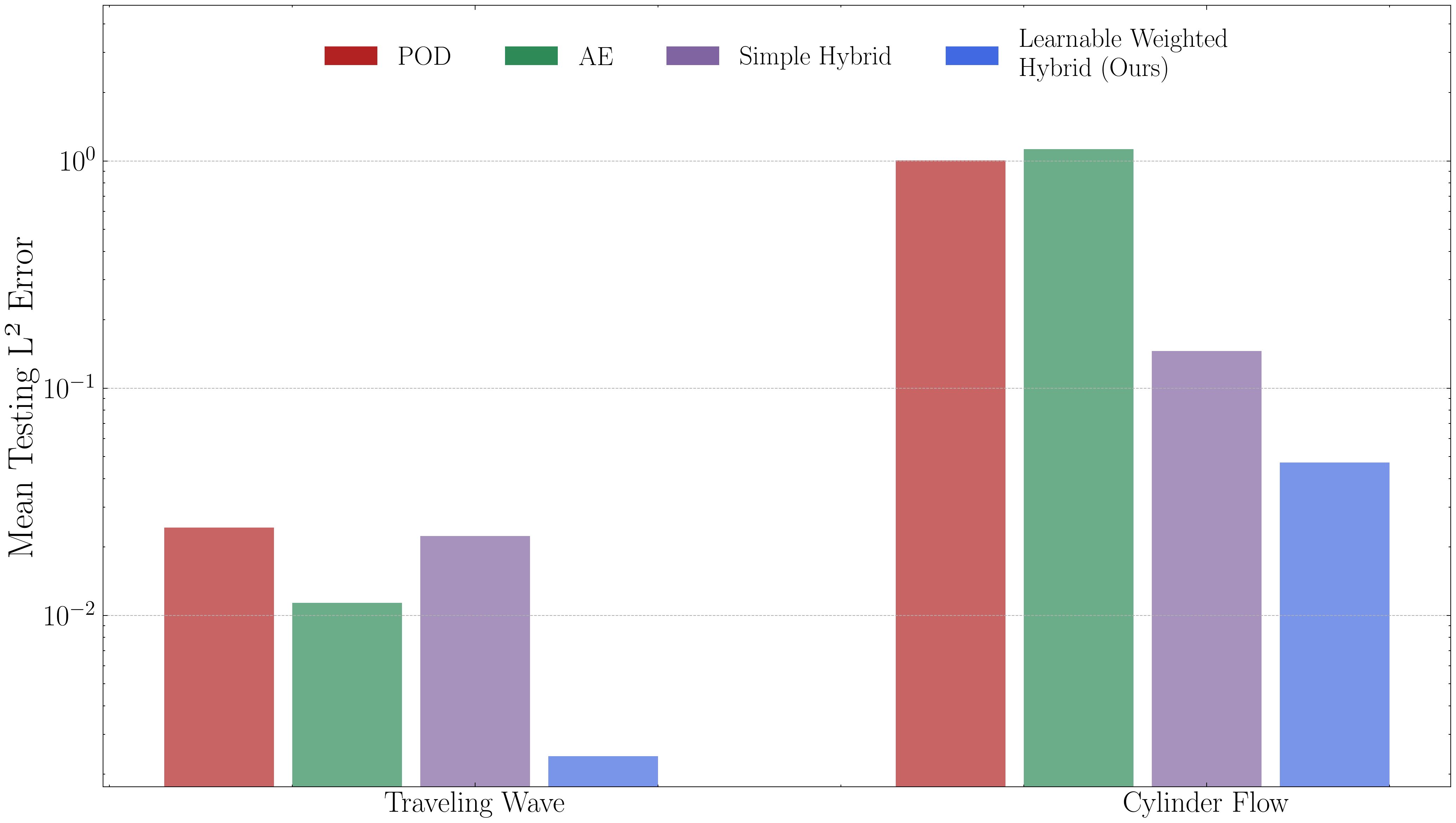}
%\rmfamily
\caption{Generalization performance of the Koopman decoder model on traveling wave and flow over cylinder dataset. Latent space rank of 2 and 4 was used for the traveling wave cylinder case respectively. Different methods for dimensionality reduction are represented by different colors. Y axis represents the Mean Testing $L^2$ error.}
\label{fig:koopman_error}
\end{figure}

\begin{figure}[h!]
\centering
%\hspace*{-0.20cm}
\includegraphics[width=1.0\textwidth]{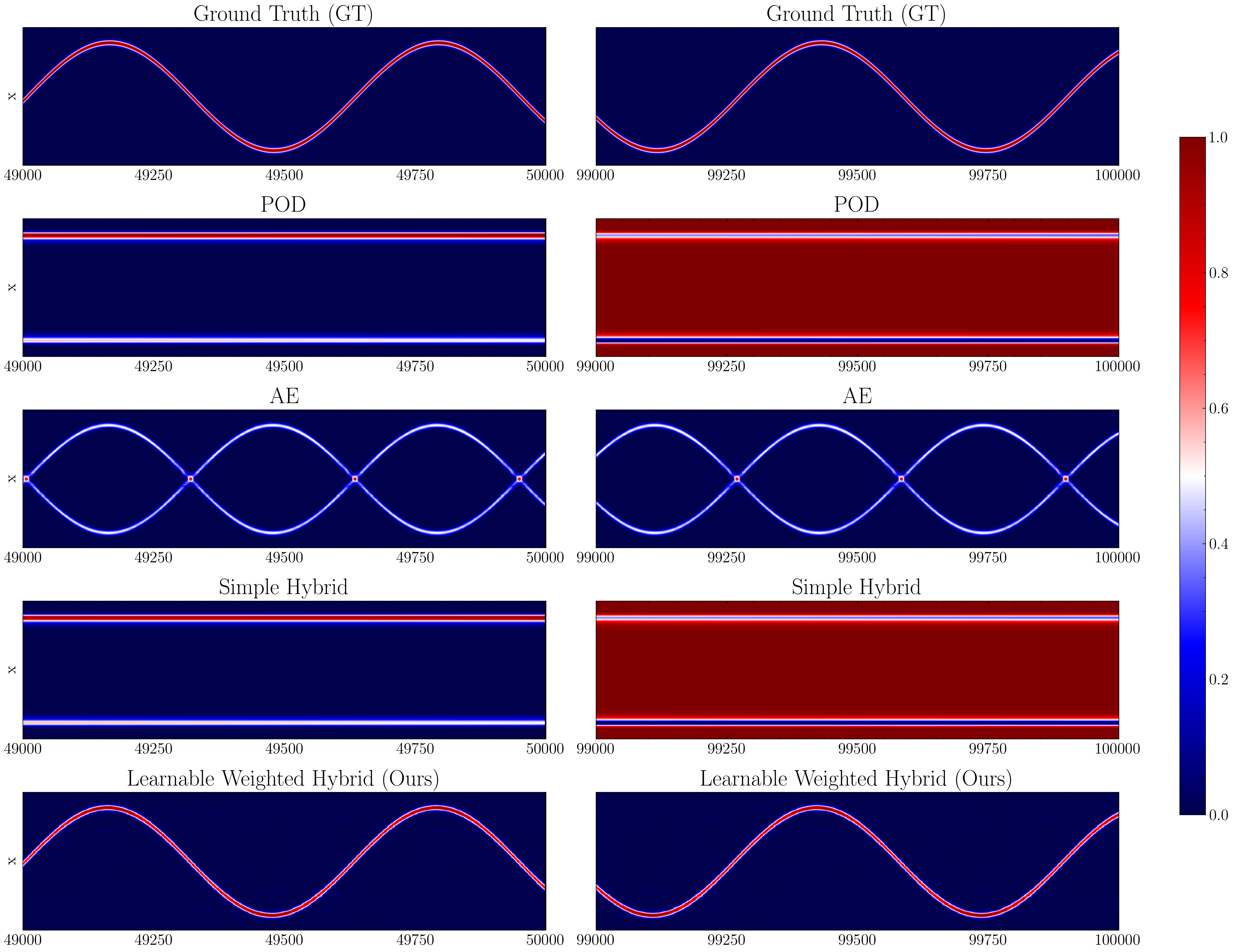}
%\rmfamily
\caption{Comparison of solution state among the different method for traveling wave for a latent space rank of 2. The training data consists of the first 50,000 timesteps and the data for the remaining 50000 timesteps are held out for testing. Left: Last 1000 snapshots within the training regime. Right: Last 1000 snapshots within the testing regime.}
\label{fig:koopman_comparison_wave}
\end{figure}

\begin{figure}[h!]
\centering
%\hspace*{-0.20cm}
\includegraphics[width=\textwidth]{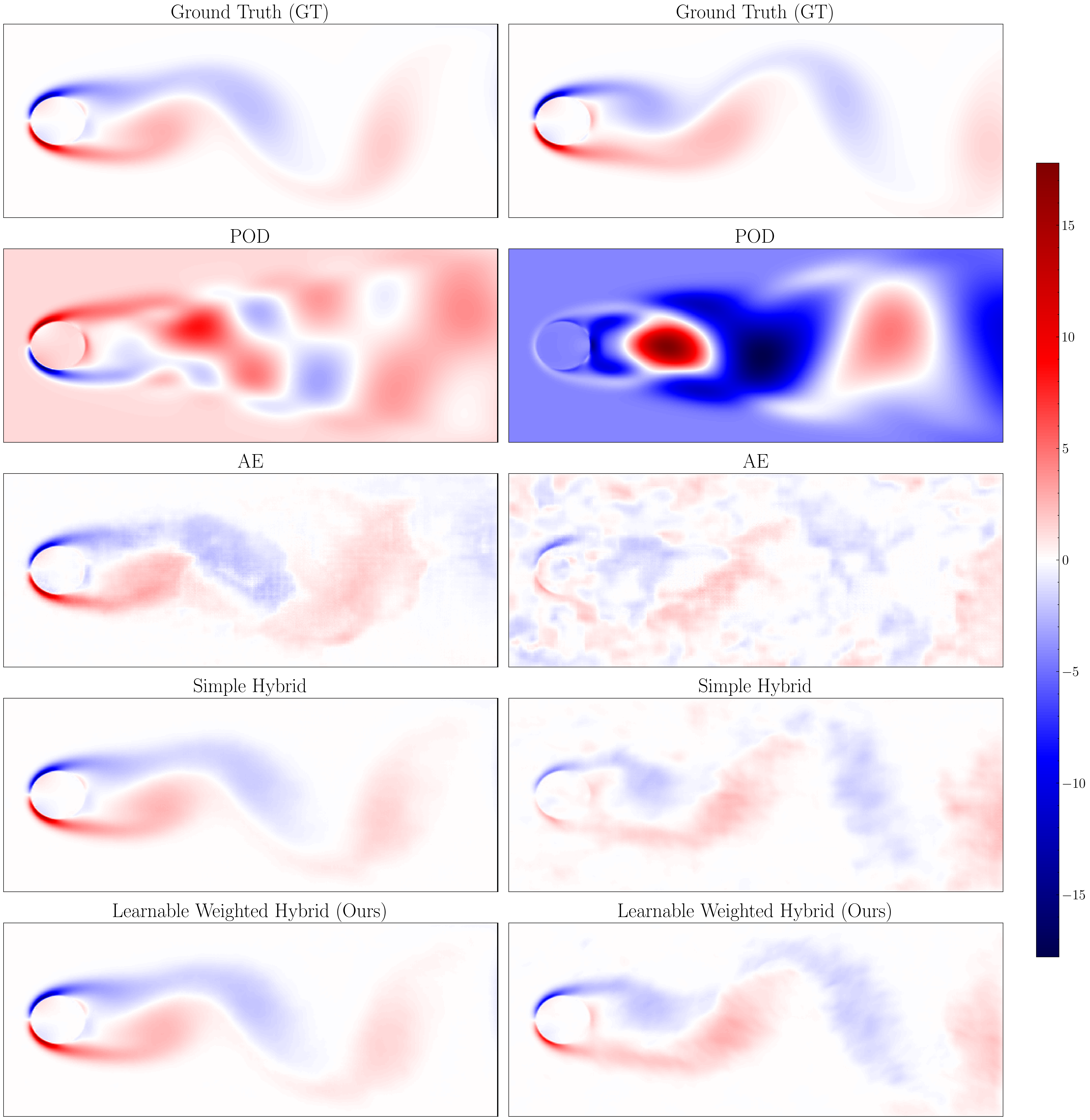}
%\rmfamily
\caption{Comparison of solution state among the different method for flow over cylinder for a latent space rank of 4. The training data consists of the first 50 timesteps and the data for the remaining 100 timesteps are held out for testing. Left: Final snapshot within the training regime (timestep 50). Right: Final snapshot within the testing regime (timestep 150).}
\label{fig:koopman_comparison_cylinder}
\end{figure}

We compare the temporal extrapolation performance of the different methods in conjunction with the Koopman forecast model. It can be seen from \Cref{fig:koopman_error}, that our approach achieves an order of magnitude lower testing error in comparison to the other methods for the traveling wave case, with simple hybrid performing similarly to POD as seen in some of the previous examples. For cylinder flow, both the simple hybrid and the proposed framework produce an order of magnitude lower testing error compared to POD and AE, and our framework further reduces the error by a factor of 2 compared to the simple hybrid. To better understand the prediction from these frameworks, \Cref{fig:koopman_comparison_wave} shows the snapshot of last 1000 timesteps of the traveling wave for the training and testing regime and the prediction from the different methods. Prediction from POD and Simple Hybrid show stationary dynamics. AE predicts a superposition of two different waves. The prediction of the proposed approach matches very well with the ground-truth data. The POD and simple hybrid approaches appear to be constrained by the expressivity of the model, preventing them from identifying a low-dimensional embedding of $r=2$ that effectively represents the system. This is a known limitation of POD for such problems. In this case, the simple hybrid was observed to be dominated by the POD component, resulting in an expressivity similar to that of POD. Although AE outperforms POD and Simple hybrid , its performance was found to be constrained by the optimization process, often getting stuck in some local minima. Meanwhile, the proposed framework effectively captures the mapping to and from the latent space and mitigates the optimization challenges, since it is initialized with POD values at the start of training. The performance of the frameworks on the training data indicates that our proposed framework effectively learns both the spatial and temporal characteristics of the dataset. This capability extends to the testing regime, demonstrating that our model generalizes well. For the cylinder flow dataset \Cref{fig:koopman_comparison_cylinder}, the proposed framework demonstrates superior generalization capabilities compared to alternative methods. Although the AE and simple hybrid approaches achieve performance comparable to our framework in the training regime, they exhibit notable degradation when the dynamics are forecast beyond the training horizon. In contrast, our model maintains high-fidelity reconstructions.  
This further emphasizes the significant improvement that can be made through the introduction of a few useful weighting parameters.   

\subsection{Surrogate Modeling Performance}
\begin{figure}[h!]
\centering
%\hspace*{-0.20cm}
\includegraphics[width=\textwidth]{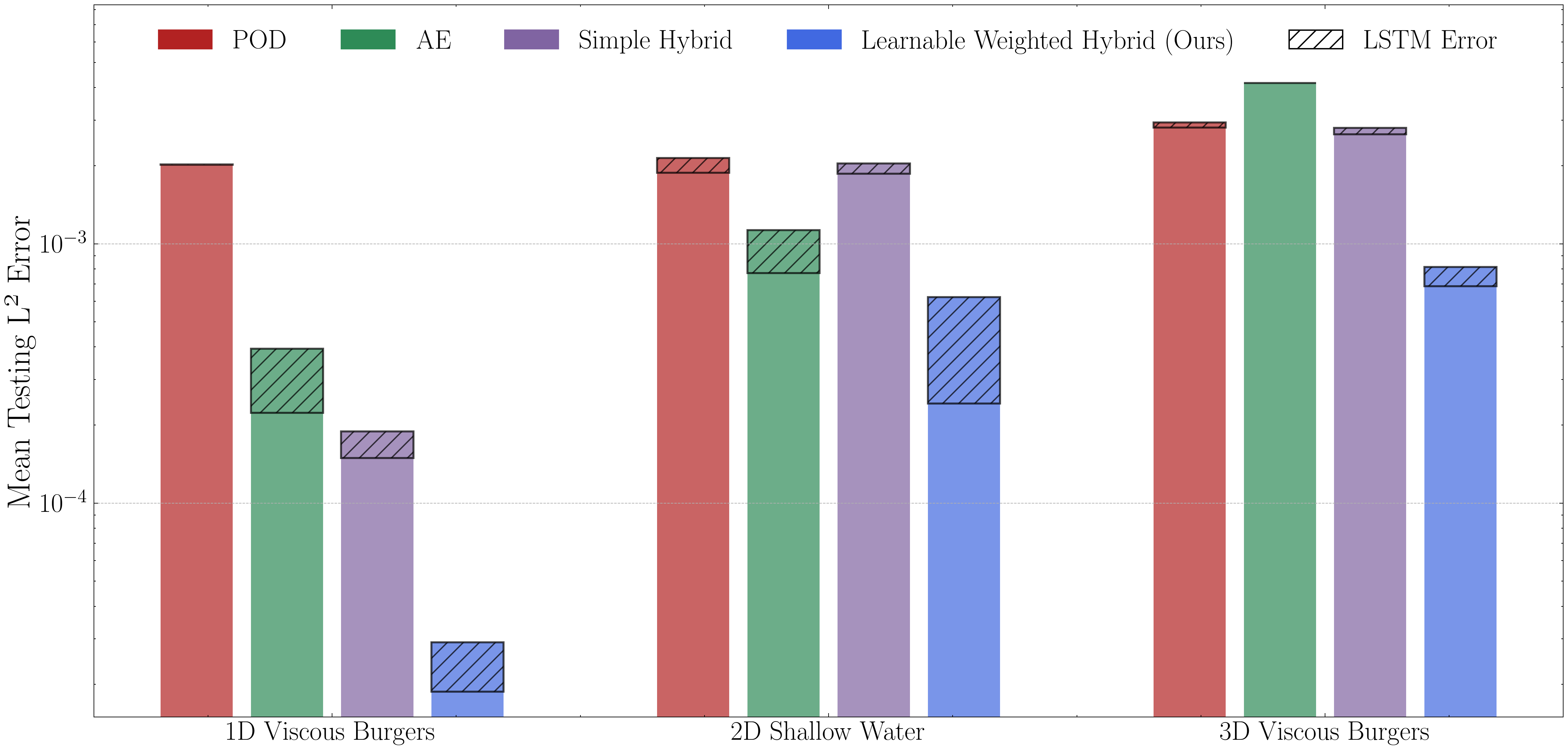}
%\rmfamily
\caption{Generalization performance of the surrogate models on the three PDEs: 1D Viscous Burgers', 2D Shallow Water and 3D Viscous Burgers'. We use a latent space rank of 2 for the 1D Viscous Burgers' case and 6 for the 2D Shallow Water and 3D Viscous Burgers' case. Different methods for dimensionality reduction are represented by different colors. Y axis represents the Mean Testing $L^2$ error. The hatched portion denotes the error contribution from the LSTM model and solid region denotes the error contribution purely from dimensionality reduction. }
\label{fig:bar_plot_surrogate}
\end{figure}

We evaluate POD, AE, Simple Hybrid, and the proposed Learnable Weighted Hybrid for time-dependent PDE surrogate modeling, where dimensionality reduction is combined with LSTM-based time series prediction to assess the impact of reduced latent representations on system dynamics forecasting. The mean test error $L^2$ for the different methods in the three PDEs considered for surrogate modeling is shown in \Cref{fig:bar_plot_surrogate}. The hatched portion denotes the error contribution from the time series modeling, and the solid region represents the error due to dimensionality reduction. In all the cases, it can be seen that our proposed approach maintains a superior performance in comparison to the other methods. We obtain almost an order-of-magnitude reduction in generalization error for the 1D Viscous Burgers. For 2D Shallow Water and 3D Viscous Burgers, the simple hybrid approach does not show much improvement over POD, while AE has the highest testing error. 

It is to be noted that error from dimensionality reduction dominates over the error introduced by system dynamics modeling by orders of magnitude. This highlights the critical importance of constructing high quality reduced representations, as any downstream application in reduced order modeling like forecasting, control, or optimization is inherently limited by the accuracy of the latent space.

\begin{figure}[h!]
\centering
%\hspace*{-0.20cm}
\includegraphics[width=\textwidth]{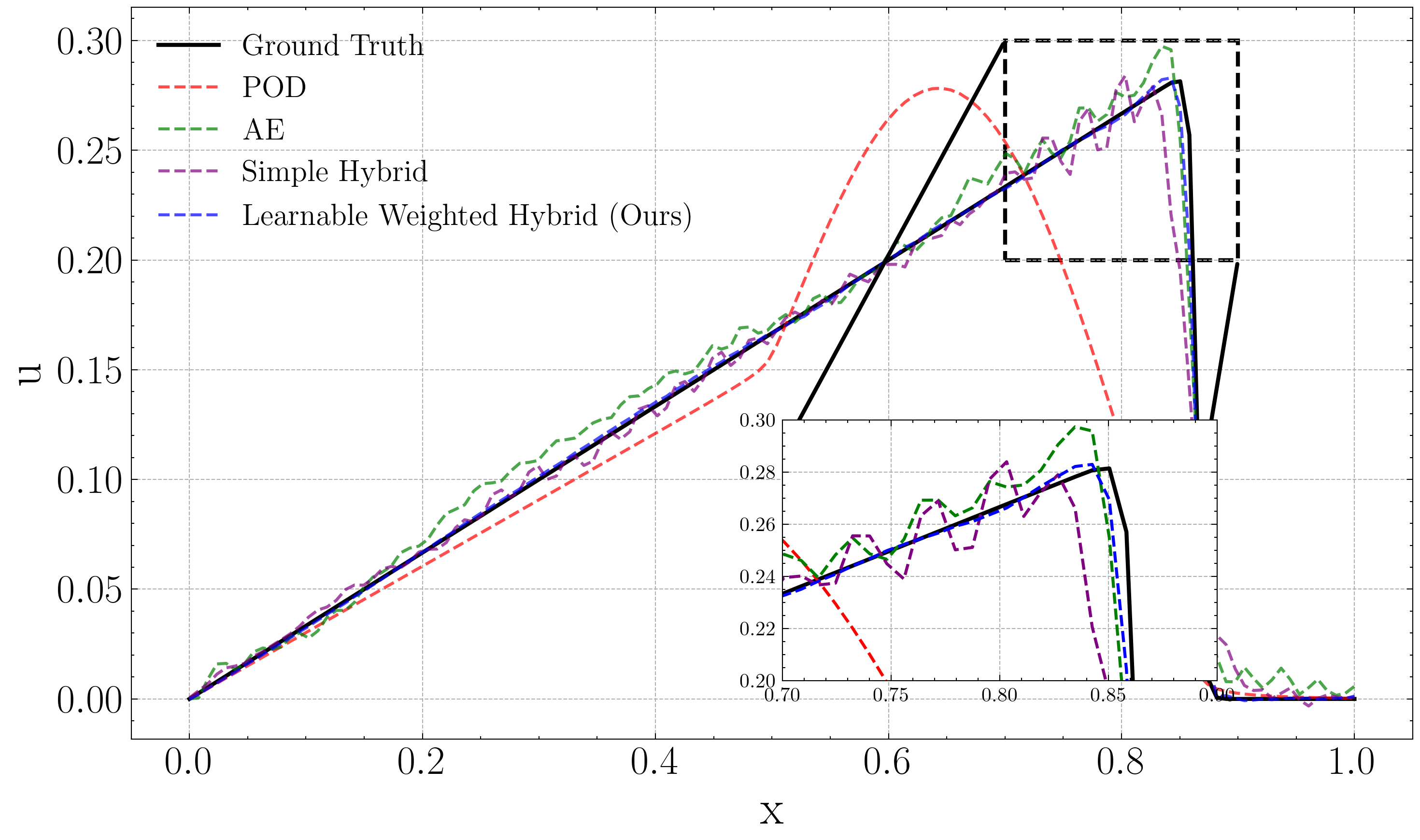}
%\rmfamily
\caption{Generalization performance comparison on 1D Viscous Burgers' data for a latent space rank of 2. The figure depicts the solution state at the final time for a trajectory with $Re=2450$. A zoomed in view near the shock location is provided for better visualization.}
\label{fig:1d_vb}
\end{figure}

To further illustrate the effectiveness of the proposed approach, we compare the reconstructed solutions for the 1D Viscous Burgers problem at the final time of 2s for a $Re$ value of 2450 in \Cref{fig:1d_vb}. In such high Reynolds number scenarios, the solution exhibits the formation and propagation of shocks over time, making accurate reduced-order modeling particularly challenging. It can be noted that methods like AE and Simple Hybrid exhibit high-frequency oscillations in space, particularly near the shock, whereas our approach maintains a smooth profile that closely conforms to the ground truth. POD solution although smooth, does not capture the location and profile of the shock accurately. This demonstrates that our method is well-suited for handling convection dominated scenarios.

\begin{figure}[h!]
\centering
%\hspace*{-0.20cm}
\includegraphics[width=\textwidth]{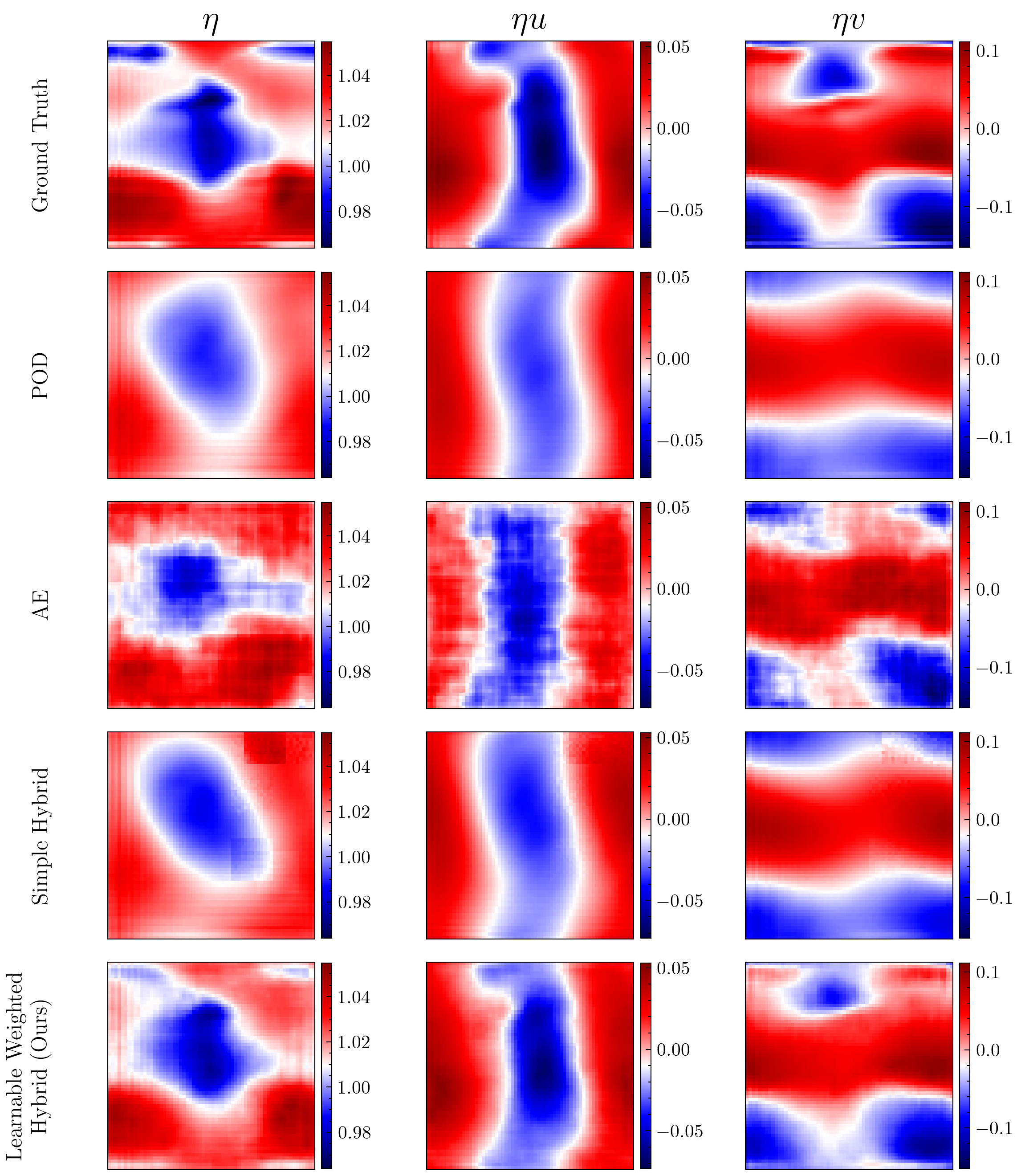}
%\rmfamily
\caption{Generalization performance comparison on 2D Shallow Water data for a latent space rank of 6. The figure depicts the solution state at the final time for a testing trajectory. Each column corresponds to a particular state variable being compared.}
\label{fig:shallow_water}
\end{figure}

\begin{figure}[h!]
\centering
%\hspace*{-0.20cm}
\includegraphics[width=\textwidth]{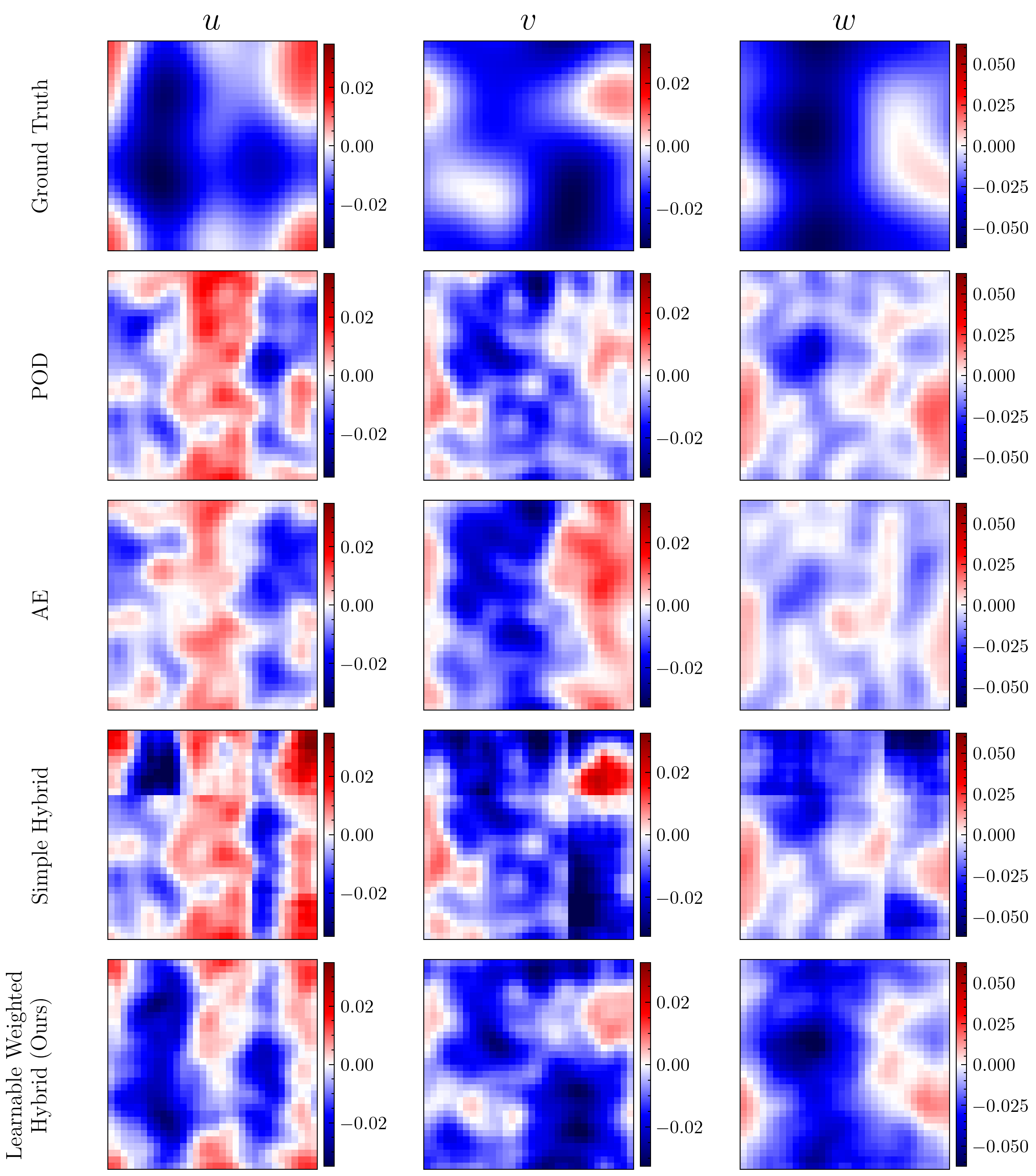}
%\rmfamily
\caption{Generalization performance comparison on 3D Viscous Burgers data for a latent space rank of 6. The figure depicts the solution state at the final time for a testing trajectory. Each column corresponds to a particular state variable being compared.}
\label{fig:3d_vb}
\end{figure}

Next, we analyze the solution at the final time of 0.5s for the 2D Shallow Water case in \Cref{fig:shallow_water}. Our approach effectively captures both large-scale and small-scale flow features within the solution domain. In contrast, POD and the Simple Hybrid approach primarily recover only the large-scale structures, while AE reconstruction contains non-sharp features. Similar comparison for the 3D Viscous Burgers can be found in \Cref{fig:3d_vb}. While the predictions from our framework do not perfectly match the ground truth, they exhibit superior physical consistency and align more closely with the true solution than other methods, despite using only six latent dimensions. The other methods tend to predict an overly positive value for the velocities and generates completely nonphysical flow features.

\subsection{Noise Robustness}

Recent studies~\cite{foret2020sharpness} in the deep learning community show that the sharpness of the minima, which describes the sensitivity of model loss with respect to perturbations in the model parameters, is a promising quantity that correlates with the generalization performance of deep networks. Let $\mathcal{D}_{\textrm{train}} = \left\{(s_1, g_1), \ldots, (s_n, g_n)\right\} $ be the training data, that is, the set of features $s$ and target $g$ pairs, and $\ell_i(\mathbf{\Theta})$ be the loss of an NN model parametrized by weights $\mathbf{\Theta}$ and evaluated at the $i$th training sample point $\left(s_i, g_i\right)$. Afterwards, the sharpness on a set of points $\mathcal{D} \subseteq \mathcal{D}_{\text {train}}$ can be defined as~\cite{andriushchenko2022towards}:
$
\mathcal{S}(\mathbf{\Theta}; \mathcal{D}) \triangleq \max_{\|\delta\|_2 \leq \rho} \frac{1}{|\mathcal{D}|} \sum_{i:\left(s_i, g_i\right) \in \mathcal{D}} (\ell_i(\mathbf{\Theta}+\delta)-\ell_i(\mathbf{\Theta}))
$ 
where $\delta$ is the perturbation introduced on the trainable parameters $\mathbf{\Theta}$ and $\rho$ refers to the perturbation radius. \textcolor{black}{For example, \Cref{tab:sharpness_ks} shows the sharpness and reconstruction error for a particular training instance of 1D KS data with grid points $1024$ and rank 60 and \Cref{tab:sharpness_hit} shows the sharpness and reconstruction error for a particular training instance of 3D HIT data with grid points $32^3$ and rank 5. The perturbation radius used here is 0.1, which is the largest value that can be applied without causing instability in the reconstructed outputs. Our proposed framework has 1000 times less sharpness compared to the AE and simple hybrid approach in the 3D HIT case}. 
Since the sharpness of the minima is related to the resilience of the model in the presence of noisy data, we evaluate the reconstruction performance of AE, a learnable weighted simple hybrid framework under noisy testing data. Models trained on noise-free data are tested on data that contain random normal noise with zero mean and standard deviation equal to $10\%$, $20\%$ and $30\%$ of the maximum velocity magnitude superimposed over the flow field. 

% \begin{table}[h!]
% \centering
% %\resizebox{0.5\textwidth}{!}{%
% \begin{tabular}{|c|c|c|}
% \hline
% Method & Sharpness &  Testing $L^2$ Error \\
% \hline
% AE & $16.69 \pm 0.9$ & 9.13 \\
% \hline
% Simple Hybrid & $22.61 \pm 0.82$ & 18.76\\
% \hline
% Learnable Weighted Hybrid (Ours) & \textbf{$0.015 \pm 0.001$} & \textbf{7.17}\\
% \hline
% \end{tabular}
% %}
% \caption{Sharpness and testing $L^2$ error for three deep autoencoders with rank $r=5$ on the 3D HIT dataset at a resolution of $32^3$.}
% \label{tab:shaprness_hit}
% \end{table}

\begin{table}[h!]
  \centering
  \resizebox{\columnwidth}{!}{
  \begin{tabular}{|c|c|p{1.9cm}|p{1.9cm}|p{1.9cm}|p{1.9cm}|}
    \hline
    \multirow{3}{*}{Method} 
      & \multirow{3}{*}{Sharpness} 
      & \multicolumn{4}{c|}{\makecell{Reconstruction $L^2$ Error Under Varying\\ Noise Level (\% of Maximum Velocity)}} \\
    \cline{3-6}
      &  & No Noise & 10\% & 20\% & 30\% \\
    \hline
    AE 
      & \makecell{\(2.21\times10^{-6}\)\\\(\pm1.42\times10^{-8}\)}
      & $5.97\times10^{-03}$ 
      & $9.27\times10^{-03}$ 
      & $4.89\times10^{-02}$ 
      & $1.43\times10^{-01}$ \\
    \hline
    Simple Hybrid 
      & \makecell{\(1.72\times10^{-6}\)\\\(\pm6.07\times10^{-7}\)}
      & $2.63\times10^{-04}$ 
      & $6.78\times10^{-04}$ 
      & $2.31\times10^{-03}$ 
      & $5.63\times10^{-02}$ \\
    \hline
    \makecell{\textbf{Learnable Weighted}\\\textbf{Hybrid (Ours)}} 
      & \makecell{\(\mathbf{8.62\times10^{-8}}\)\\\(\pm\mathbf{9.47\times10^{-10}}\)}
      & $\mathbf{2.56\times10^{-05}}$ 
      & $\mathbf{3.12\times10^{-05}}$ 
      & $\mathbf{6.78\times10^{-05}}$ 
      & $\mathbf{1.18\times10^{-04}}$ \\
    \hline
  \end{tabular}}
  \caption{\textcolor{black}{Sharpness and reconstruction $L^2$ error for three deep autoencoders with rank $r=60$ on the 1D KS dataset at a resolution of $1024$ under varying noise levels in the test input.}}
  \label{tab:sharpness_ks}
\end{table}

\begin{table}[h!]
\centering
\resizebox{\columnwidth}{!}{
\begin{tabular}{|c|c|p{1.5cm}|p{1.5cm}|p{1.5cm}|p{1.5cm}|}
\hline
\multirow{3}{*}{Method} & \multirow{3}{*}{Sharpness} & \multicolumn{4}{c|}{\makecell{Reconstruction $L^2$ Error Under Varying \\ Noise Level (\% of Maximum Velocity)}} \\
\cline{3-6}
 & & No Noise & 10\% & 20\% & 30\% \\
\hline
AE & $16.69 \pm 0.9$ & ${9.13}$ & 11.50 & 25.95 & 41.20 \\
\hline
Simple Hybrid & $22.61 \pm 0.82$ & ${18.76}$ & 22.26 & 36.93 & 58.12 \\
\hline
\makecell{\textbf{Learnable Weighted} \\ \textbf{Hybrid (Ours)}} & $\mathbf{0.015} \pm \mathbf{0.001}$ & $\mathbf{7.17}$ & $\mathbf{8.12}$ & $\mathbf{9.33}$ & $\mathbf{9.43}$ \\
\hline
\end{tabular}}
\caption{Sharpness and reconstruction $L^2$ error for three deep autoencoders with rank $r=5$ on the 3D HIT dataset at a resolution of $32^3$ under varying noise levels in the test input.}
\label{tab:sharpness_hit}
\end{table}

% \begin{figure}[h!]
% \centering
% %\hspace*{-0.5cm}
% \includegraphics[width=0.8\textwidth]{./figs/3d_hit/error_vs_noise.png}
% %\rmfamily
% \caption{Comparison of testing $L^2$ error for varying noise level for three different deep autoencoder frameworks on 3D HIT dataset at resolution $32^3$ and rank $r=5$. }
% \label{fig:error_vs_noise}
% \end{figure}

\begin{figure}[h!]
\centering
%\hspace*{-0.2cm}
\includegraphics[width=\textwidth]{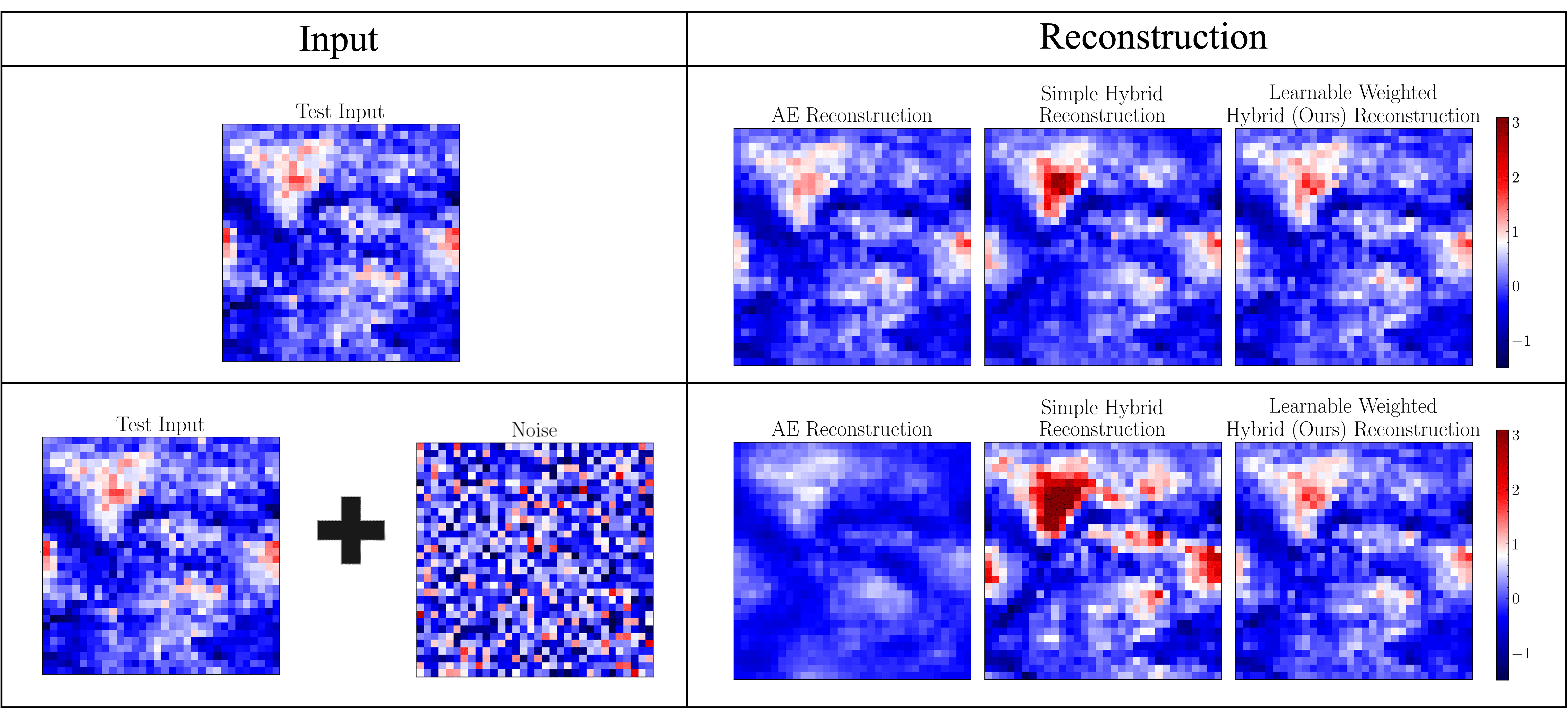}
%\rmfamily
\caption{Comparison of the reconstruction performance of three different deep autoencoder frameworks on 3D HIT dataset at resolution $32^3$ and rank $r=5$. Up: noise free. Down: 30\% noise level.}
\label{fig:noise_30}
\end{figure}

The reconstruction error for all models is shown in the same table. As the noise level increases, there is a drastic increase in the testing error for both the AE and the simple hybrid method, while our approach stays the same,  demonstrating the robustness of this framework against noise in the unseen test data. \textcolor{black}{For the 3D HIT case,} a 2D slice of $u_y$ with and without 30\% noise level is further visualized in \Cref{fig:noise_30}, which shows the improved robustness of our approach over other methods.

% The improved generalization performance of our proposed approach can be attributed to solving the optimization problem at a better starting point. The tunable approach effectively uses POD results as the starting point of the minimization problem, while the other approaches uses random guess. 

% Another aspect to consider here is the sharpness \cite{andriushchenko2022understandingsharpnessawareminimization} of the optima that is being computed. 

% \begin{figure}[ht]
% \centering
% %\hspace*{-0.5cm}
% \includegraphics[width=0.8\textwidth]{./figs/surrogate_modeling/CAE_LSTM.png}
% %\rmfamily
% \caption{Architecture of surrogate modeling framework. The dimensionality reduction framework is trained and using its latent space predictions, we train the LSTM for temporal evolution of the system.}
% \label{fig:cae_lstm}
% \end{figure}

\section{Conclusions}\label{sec::conc}
In this work, we present a novel deep autoencoder framework that demonstrates convergence properties akin to SVD. By incorporating a learnable weighted average between SVD and vanilla deep autoencoders (either feedforward or convolutional), our approach achieves SVD-like convergence as the rank increases. We validate the effectiveness of this framework on pure reconstruction tasks using two challenging chaotic PDE datasets: the 1D Kuramoto-Sivashinsky and the 3D homogeneous isotropic turbulence. The results show that our learnable weighted hybrid autoencoder consistently achieves the lowest testing error and exhibits superior robustness to noisy data compared to other methods such as POD, vanilla deep autoencoders, and simple hybrid autoencoders. Remarkably, we find that our proposed approach leads to a minimum with a sharpness that is a thousand times smaller than that of other deep autoencoder frameworks. In addition, we demonstrate that utilizing our proposed framework in tandem with time series prediction models, we can achieve superior performance for surrogate modeling of time-dependent PDEs over other approaches. Our framework is also capable of capturing dynamics of the system with strong discontinuities without spurious oscillations in the solution. This highlights its potential for robust and generalizable representation learning in complex PDE systems.\vskip6pt

\FloatBarrier
\section*{CRediT authorship contribution statement}

\textbf{Nithin Somasekharan}: Data Curation (lead), Formal
Analysis (lead), Investigation (lead), Software (lead), Visualization (lead), Writing – Original Draft Preparation (lead). \textbf{Shaowu Pan}: Conceptualization (lead), Funding Acquisition (lead), Methodology (lead), Supervision (lead), Writing – Review \& Editing (lead), Project Administration (lead).

\enlargethispage{20pt}

\section*{Acknowledgement}
{This work was supported by U.S. Department of Energy under Advancements in Artificial Intelligence for Science with award number DE-SC0025425. 
The authors thank the Center for Computational Innovations (CCI) at Rensselaer Polytechnic Institute (RPI) for providing computational resources during the early stages of this research. Numerical experiments are performed using computational resources granted by NSF-ACCESS for the project PHY240112 and that of the National Energy Research Scientific Computing Center, a DOE Office of Science User Facility using the NERSC award NERSC DDR-ERCAP0030714.}

\appendix

\renewcommand{\thetable}{\arabic{table}}
% table for KS
\section{Detailed comparison of numerical results}
We summarize the detailed results of all numerical experiments conducted in this work in \Cref{tab:ks_table,tab:hit_table}. Upon publication, the code and data will be available at \url{https://github.com/csml-rpi/deep-ae-with-svd-convergence}.

\begin{table*}[htbp]
\centering
\rmfamily
\resizebox{\textwidth}{!}{
\begin{tabular}{rrcccccccccccc}
\toprule
Grid & Rank & \multicolumn{4}{c}{Mean Train Error, $\mu$($\pm \sigma$)} & \multicolumn{4}{c}{Mean Test Error, $\mu$($\pm \sigma$)} & \multicolumn{4}{c}{Total Number of Model Parameters} \\
\cmidrule(lr){3-6} \cmidrule(lr){7-10} \cmidrule(lr){11-14}
& & POD & AE & {Simple} & \textbf{Learnable Weighted} & POD & AE & {Simple} & \textbf{Learnable Weighted} & POD & AE & {Simple} & \textbf{Learnable Weighted} \\
& & & & {Hybrid} & \textbf{Hybrid (ours)} & & & {Hybrid} & \textbf{Hybrid (ours)} & & & {Hybrid} & \textbf{Hybrid (ours)} \\
\midrule
\multirow{10}{*}{512} 
  & 50 
    & $2.84\times10^{-1}$ & $2.54\times10^{-3}$ & $2.41\times10^{-4}$ & \bm{$1.62\times10^{-4}$}
    & $3.06\times10^{-1}$ & $2.73\times10^{-3}$ & $2.55\times10^{-4}$ & \bm{$1.92\times10^{-4}$}
    & 25,600 & 113,162 & 138,762 & 138,813 \\ 
  & 
    & ($\pm1.92\times10^{-3}$) & ($\pm1.42\times10^{-3}$) & ($\pm6.07\times10^{-5}$) & (\bm{$\pm9.47\times10^{-6}$})
    & ($\pm4.80\times10^{-3}$) & ($\pm1.49\times10^{-3}$) & ($\pm5.96\times10^{-5}$) & (\bm{$\pm1.03\times10^{-5}$})
    &  &  &  &  \\
\addlinespace
  & 60 
    & $6.89\times10^{-2}$ & $1.68\times10^{-3}$ & $1.88\times10^{-4}$ & \bm{$1.59\times10^{-5}$}
    & $7.50\times10^{-2}$ & $1.76\times10^{-3}$ & $1.94\times10^{-4}$ & \bm{$1.96\times10^{-5}$}
    & 30,720 & 138,092 & 168,812 & 168,873 \\ 
  & 
    & ($\pm4.92\times10^{-4}$) & ($\pm5.84\times10^{-4}$) & ($\pm7.35\times10^{-5}$) & (\bm{$\pm7.32\times10^{-7}$})
    & ($\pm1.47\times10^{-3}$) & ($\pm5.75\times10^{-4}$) & ($\pm7.38\times10^{-5}$) & (\bm{$\pm2.94\times10^{-6}$})
    &  &  &  &  \\
\addlinespace
  & 70 
    & $1.83\times10^{-2}$ & $1.75\times10^{-3}$ & $1.93\times10^{-4}$ & \bm{$2.73\times10^{-6}$}
    & $2.03\times10^{-2}$ & $1.79\times10^{-3}$ & $1.97\times10^{-4}$ & \bm{$3.55\times10^{-6}$}
    & 35,840 & 163,822 & 199,662 & 199,733 \\ 
  & 
    & ($\pm1.37\times10^{-4}$) & ($\pm1.14\times10^{-3}$) & ($\pm1.26\times10^{-4}$) & (\bm{$\pm3.95\times10^{-7}$})
    & ($\pm3.67\times10^{-4}$) & ($\pm1.10\times10^{-3}$) & ($\pm1.28\times10^{-4}$) & (\bm{$\pm9.03\times10^{-7}$})
    &  &  &  &  \\
\addlinespace
  & 80 
    & $4.59\times10^{-3}$ & $2.07\times10^{-3}$ & $1.55\times10^{-4}$ & \bm{$1.02\times10^{-6}$}
    & $5.16\times10^{-3}$ & $2.22\times10^{-3}$ & $1.58\times10^{-4}$ & \bm{$1.39\times10^{-6}$}
    & 40,960 & 190,352 & 231,312 & 231,393 \\ 
  & 
    & ($\pm2.95\times10^{-5}$) & ($\pm1.68\times10^{-3}$) & ($\pm4.62\times10^{-5}$) & (\bm{$\pm1.05\times10^{-7}$})
    & ($\pm8.78\times10^{-5}$) & ($\pm1.75\times10^{-3}$) & ($\pm4.72\times10^{-5}$) & (\bm{$\pm5.40\times10^{-7}$})
    &  &  &  &  \\
\midrule
\multirow{10}{*}{1024} 
  & 50 
    & $3.18\times10^{-1}$ & $6.48\times10^{-3}$ & $3.01\times10^{-4}$ & \bm{$2.53\times10^{-4}$}
    & $3.41\times10^{-1}$ & $6.95\times10^{-3}$ & $3.20\times10^{-4}$ & \bm{$2.98\times10^{-4}$}
    & 51,200 & 216,074 & 267,274 & 267,325 \\ 
  & 
    & ($\pm2.28\times10^{-3}$) & ($\pm2.53\times10^{-3}$) & ($\pm6.88\times10^{-5}$) & (\bm{$\pm2.29\times10^{-5}$})
    & ($\pm5.12\times10^{-3}$) & ($\pm2.49\times10^{-3}$) & ($\pm6.89\times10^{-5}$) & (\bm{$\pm2.29\times10^{-5}$})
    &  &  &  &  \\
\addlinespace
  & 60 
    & $8.75\times10^{-2}$ & $6.10\times10^{-3}$ & $2.14\times10^{-4}$ & \bm{$2.89\times10^{-5}$}
    & $9.53\times10^{-2}$ & $6.19\times10^{-3}$ & $2.22\times10^{-4}$ & \bm{$3.61\times10^{-5}$}
    & 61,440 & 261,484 & 322,924 & 322,985 \\ 
  & 
    & ($\pm4.46\times10^{-4}$) & ($\pm7.01\times10^{-3}$) & ($\pm6.56\times10^{-5}$) & (\bm{$\pm3.29\times10^{-6}$})
    & ($\pm1.11\times10^{-3}$) & ($\pm6.43\times10^{-3}$) & ($\pm6.54\times10^{-5}$) & (\bm{$\pm5.84\times10^{-6}$})
    &  &  &  &  \\
\addlinespace
  & 70 
    & $2.21\times10^{-2}$ & $3.76\times10^{-3}$ & $2.25\times10^{-4}$ & \bm{$5.33\times10^{-6}$}
    & $2.46\times10^{-2}$ & $4.08\times10^{-3}$ & $2.30\times10^{-4}$ & \bm{$7.10\times10^{-6}$}
    & 71,680 & 307,694 & 379,374 & 379,445 \\ 
  & 
    & ($\pm9.31\times10^{-5}$) & ($\pm2.64\times10^{-3}$) & ($\pm1.31\times10^{-4}$) & (\bm{$\pm1.31\times10^{-6}$})
    & ($\pm3.32\times10^{-4}$) & ($\pm2.72\times10^{-3}$) & ($\pm1.30\times10^{-4}$) & (\bm{$\pm2.38\times10^{-6}$})
    &  &  &  &  \\
\addlinespace
  & 80 
    & $5.86\times10^{-3}$ & $3.69\times10^{-3}$ & $2.23\times10^{-4}$ & \bm{$1.94\times10^{-6}$}
    & $6.58\times10^{-3}$ & $4.00\times10^{-3}$ & $2.26\times10^{-4}$ & \bm{$2.68\times10^{-6}$}
    & 81,920 & 354,704 & 436,624 & 436,705 \\ 
  & 
    & ($\pm3.16\times10^{-5}$) & ($\pm2.34\times10^{-3}$) & ($\pm1.07\times10^{-4}$) & (\bm{$\pm1.54\times10^{-7}$})
    & ($\pm9.69\times10^{-5}$) & ($\pm2.34\times10^{-3}$) & ($\pm1.06\times10^{-4}$) & (\bm{$\pm8.49\times10^{-7}$})
    &  &  &  &  \\
\midrule
\multirow{10}{*}{2048} 
  & 50 
    & $2.92\times10^{-1}$ & $1.62\times10^{-2}$ & $2.74\times10^{-4}$ & \bm{$2.36\times10^{-4}$}
    & $3.15\times10^{-1}$ & $1.82\times10^{-2}$ & $2.98\times10^{-4}$ & \bm{$2.80\times10^{-4}$}
    & 102,400 & 421,898 & 524,298 & 524,349 \\ 
  & 
    & ($\pm2.34\times10^{-3}$) & ($\pm1.81\times10^{-2}$) & ($\pm7.71\times10^{-5}$) & (\bm{$\pm2.03\times10^{-5}$})
    & ($\pm5.32\times10^{-3}$) & ($\pm1.88\times10^{-2}$) & ($\pm8.97\times10^{-5}$) & (\bm{$\pm3.50\times10^{-5}$})
    &  &  &  &  \\
\addlinespace
  & 60 
    & $7.20\times10^{-2}$ & $9.93\times10^{-3}$ & $1.65\times10^{-4}$ & \bm{$3.39\times10^{-5}$}
    & $7.84\times10^{-2}$ & $1.14\times10^{-2}$ & $1.76\times10^{-4}$ & \bm{$4.35\times10^{-5}$}
    & 122,880 & 508,268 & 631,148 & 631,209 \\ 
  & 
    & ($\pm4.95\times10^{-4}$) & ($\pm4.29\times10^{-3}$) & ($\pm2.48\times10^{-5}$) & (\bm{$\pm2.63\times10^{-6}$})
    & ($\pm1.31\times10^{-3}$) & ($\pm4.79\times10^{-3}$) & ($\pm2.31\times10^{-5}$) & (\bm{$\pm1.05\times10^{-5}$})
    &  &  &  &  \\
\addlinespace
  & 70 
    & $1.80\times10^{-2}$ & $7.65\times10^{-3}$ & $2.52\times10^{-4}$ & \bm{$1.21\times10^{-5}$}
    & $1.99\times10^{-2}$ & $9.18\times10^{-3}$ & $2.59\times10^{-4}$ & \bm{$1.61\times10^{-5}$}
    & 143,360 & 595,438 & 738,798 & 738,869 \\ 
  & 
    & ($\pm9.85\times10^{-5}$) & ($\pm3.15\times10^{-3}$) & ($\pm7.35\times10^{-5}$) & (\bm{$\pm6.28\times10^{-7}$})
    & ($\pm3.20\times10^{-4}$) & ($\pm3.47\times10^{-3}$) & ($\pm7.46\times10^{-5}$) & (\bm{$\pm4.49\times10^{-6}$})
    &  &  &  &  \\
\addlinespace
  & 80 
    & $4.72\times10^{-3}$ & $9.62\times10^{-3}$ & $3.07\times10^{-4}$ & \bm{$9.23\times10^{-6}$}
    & $5.34\times10^{-3}$ & $1.12\times10^{-2}$ & $3.13\times10^{-4}$ & \bm{$1.23\times10^{-5}$}
    & 163,840 & 683,408 & 847,248 & 847,329 \\ 
  & 
    & ($\pm3.99\times10^{-5}$) & ($\pm6.61\times10^{-3}$) & ($\pm2.34\times10^{-4}$) & (\bm{$\pm3.11\times10^{-7}$})
    & ($\pm1.54\times10^{-4}$) & ($\pm6.77\times10^{-3}$) & ($\pm2.30\times10^{-4}$) & (\bm{$\pm2.95\times10^{-6}$})
    &  &  &  &  \\
\bottomrule
\end{tabular}
}
\caption{Summary of training $L^2$ error, testing $L^2$ error, and the number of parameters for the four models trained on data with varying grid resolutions and ranks $r$ pertaining to 1D K-S case. The standard deviation of the error is indicated in parentheses. The number of parameters for the hybrid approaches includes non-trainable POD parameters, which remain fixed throughout the optimization process and are not trainable.}
\label{tab:ks_table}
\end{table*}

% -- Second table --

\begin{table*}[htbp]
\centering
\rmfamily
\resizebox{\textwidth}{!}{
\begin{tabular}{rrcccccccccccc}
\toprule
Grid & Rank & \multicolumn{4}{c}{Mean Train Error, $\mu$($\pm \sigma$)} & \multicolumn{4}{c}{Mean Test Error, $\mu$($\pm \sigma$)} & \multicolumn{4}{c}{Total Number of Model Parameters} \\
\cmidrule(lr){3-6} \cmidrule(lr){7-10} \cmidrule(lr){11-14}
& & POD & AE & {Simple} & \textbf{Learnable Weighted} & POD & AE & {Simple} & \textbf{Learnable Weighted} & POD & AE & {Simple} & \textbf{Learnable Weighted} \\
& & & & {Hybrid} & \textbf{Hybrid (ours)} & & & {Hybrid} & \textbf{Hybrid (ours)} & & & {Hybrid} & \textbf{Hybrid (ours)} \\
\midrule
\multirow{10}{*}{$16^3$} 
  & 5 
    & $2.01\times10^{1}$ & \bm{$1.05\times10^{0}$} & $1.35\times10^{0}$ & $1.76\times10^{0}$
    & $2.04\times10^{1}$ & $5.74\times10^{0}$ & $5.98\times10^{0}$ & \bm{$5.48\times10^{0}$}
    & 61,440 & 688,348,168 & 688,409,608 & 688,409,616 \\ 
  & 
    & ($\pm0.00$) & (\bm{$\pm1.37\times10^{-1}$}) & ($\pm1.13\times10^{-1}$) & ($\pm9.31\times10^{-2}$)
    & ($\pm0.00$) & ($\pm3.13\times10^{-1}$) & ($\pm1.09\times10^{-1}$) & (\bm{$\pm6.44\times10^{-2}$})
    &  &  &  &  \\
\addlinespace
  & 10 
    & $1.41\times10^{1}$ & $9.71\times10^{-1}$ & \bm{$6.85\times10^{-1}$} & $9.02\times10^{-1}$
    & $1.47\times10^{1}$ & $5.53\times10^{0}$ & \bm{$5.17\times10^{0}$} & $5.22\times10^{0}$
    & 122,880 & 688,368,653 & 688,491,533 & 688,491,546 \\ 
  & 
    & ($\pm0.0$) & ($\pm2.75\times10^{-1}$) & (\bm{$\pm6.05\times10^{-2}$}) & ($\pm4.04\times10^{-2}$)
    & ($\pm0.0$) & ($\pm1.11\times10^{-1}$) & (\bm{$\pm3.20\times10^{-2}$}) & ($\pm3.49\times10^{-2}$)
    &  &  &  &  \\
\addlinespace
  & 15 
    & $1.14\times10^{1}$ & $8.76\times10^{-1}$ & \bm{$4.65\times10^{-1}$} & $6.35\times10^{-1}$
    & $1.23\times10^{1}$ & $5.47\times10^{0}$ & \bm{$5.08\times10^{0}$} & $5.20\times10^{0}$
    & 184,320 & 688,389,138 & 688,573,458 & 688,573,476 \\ 
  & 
    & ($\pm0.0$) & ($\pm4.36\times10^{-2}$) & (\bm{$\pm2.43\times10^{-2}$}) & ($\pm3.55\times10^{-2}$)
    & ($\pm0.00$) & ($\pm1.02\times10^{-1}$) & (\bm{$\pm3.22\times10^{-2}$}) & ($\pm3.59\times10^{-2}$)
    &  &  &  &  \\
\addlinespace
  & 20 
    & $9.71\times10^{0}$ & $8.96\times10^{-1}$ & \bm{$3.07\times10^{-1}$} & $5.06\times10^{-1}$
    & $1.09\times10^{1}$ & $5.49\times10^{0}$ & \bm{$4.85\times10^{0}$} & $5.21\times10^{0}$
    & 245,760 & 688,409,623 & 688,655,383 & 688,655,406 \\ 
  & 
    & ($\pm0.00$) & ($\pm5.07\times10^{-2}$) & (\bm{$\pm1.90\times10^{-2}$}) & ($\pm2.83\times10^{-2}$)
    & ($\pm0.00$) & ($\pm6.92\times10^{-2}$) & (\bm{$\pm3.01\times10^{-2}$}) & ($\pm3.14\times10^{-2}$)
    &  &  &  &  \\
\midrule
\multirow{10}{*}{$32^3$} 
  & 5 
    & $2.02\times10^{1}$ & $9.15\times10^{0}$ & $1.84\times10^{1}$ & \bm{$5.56\times10^{0}$}
    & $2.04\times10^{1}$ & $1.05\times10^{1}$ & $1.87\times10^{1}$ & \bm{$7.16\times10^{0}$}
    & 491,520 & 688,505,864 & 688,997,384 & 688,997,392 \\ 
  & 
    & ($\pm0.00$) & ($\pm1.10\times10^{1}$) & ($\pm2.61\times10^{-1}$) & (\bm{$\pm1.03\times10^{-1}$})
    & ($\pm0.00$) & ($\pm1.05\times10^{1}$) & ($\pm2.34\times10^{-1}$) & (\bm{$\pm7.28\times10^{-2}$})
    &  &  &  &  \\
\addlinespace
  & 10 
    & $1.42\times10^{1}$ & $1.52\times10^{1}$ & $1.23\times10^{1}$ & \bm{$3.98\times10^{0}$}
    & $1.47\times10^{1}$ & $1.63\times10^{1}$ & $1.32\times10^{1}$ & \bm{$6.50\times10^{0}$}
    & 983,040 & 688,669,709 & 689,652,749 & 689,652,762 \\ 
  & 
    & ($\pm0.00$) & ($\pm1.74\times10^{1}$) & ($\pm1.45\times10^{-1}$) & (\bm{$\pm6.67\times10^{-2}$})
    & ($\pm0.00$) & ($\pm1.66\times10^{1}$) & ($\pm1.27\times10^{-1}$) & (\bm{$\pm5.38\times10^{-2}$})
    &  &  &  &  \\
\addlinespace
  & 15 
    & $1.15\times10^{1}$ & $1.46\times10^{1}$ & $9.18\times10^{0}$ & \bm{$2.90\times10^{0}$}
    & $1.22\times10^{1}$ & $1.59\times10^{1}$ & $1.06\times10^{1}$ & \bm{$6.18\times10^{0}$}
    & 1,474,560 & 688,833,554 & 690,308,114 & 690,308,132 \\ 
  & 
    & ($\pm0.00$) & ($\pm1.76\times10^{1}$) & ($\pm1.81\times10^{-1}$) & (\bm{$\pm5.54\times10^{-2}$})
    & ($\pm0.00$) & ($\pm1.68\times10^{1}$) & ($\pm1.55\times10^{-1}$) & (\bm{$\pm4.65\times10^{-2}$})
    &  &  &  &  \\
\addlinespace
  & 20 
    & $9.78\times10^{0}$ & $1.10\times10^{1}$ & $6.91\times10^{0}$ & \bm{$2.21\times10^{0}$}
    & $1.08\times10^{1}$ & $1.26\times10^{1}$ & $8.86\times10^{0}$ & \bm{$5.97\times10^{0}$}
    & 1,966,080 & 688,997,399 & 690,963,479 & 690,963,502 \\ 
  & 
    & ($\pm0.00$) & ($\pm1.52\times10^{1}$) & ($\pm2.84\times10^{-1}$) & (\bm{$\pm4.70\times10^{-2}$})
    & ($\pm0.00$) & ($\pm1.44\times10^{1}$) & ($\pm1.55\times10^{-1}$) & (\bm{$\pm3.96\times10^{-2}$})
    &  &  &  &  \\
\midrule
\multirow{10}{*}{$64^3$} 
  & 5 
    & $2.02\times10^{1}$ & $2.89\times10^{1}$ & $1.95\times10^{1}$ & \bm{$9.00\times10^{0}$}
    & $2.04\times10^{1}$ & $2.89\times10^{1}$ & $1.96\times10^{1}$ & \bm{$9.53\times10^{0}$}
    & 3,932,160 & 689,767,432 & 693,699,592 & 693,699,600 \\ 
  & 
    & ($\pm0.00$) & ($\pm1.94\times10^{1}$) & ($\pm1.53\times10^{-1}$) & (\bm{$\pm9.62\times10^{-2}$})
    & ($\pm0.00$) & ($\pm1.91\times10^{1}$) & ($\pm1.48\times10^{-1}$) & (\bm{$\pm9.27\times10^{-2}$})
    &  &  &  &  \\
\addlinespace
  & 10 
    & $1.42\times10^{1}$ & $2.50\times10^{1}$ & $1.35\times10^{1}$ & \bm{$7.01\times10^{0}$}
    & $1.46\times10^{1}$ & $2.52\times10^{1}$ & $1.40\times10^{1}$ & \bm{$7.95\times10^{0}$}
    & 7,864,320 & 691,078,157 & 698,942,477 & 698,942,490 \\ 
  & 
    & ($\pm0.00$) & ($\pm2.00\times10^{1}$) & ($\pm1.22\times10^{-1}$) & (\bm{$\pm7.07\times10^{-2}$})
    & ($\pm0.00$) & ($\pm1.96\times10^{1}$) & ($\pm1.18\times10^{-1}$) & (\bm{$\pm6.18\times10^{-2}$})
    &  &  &  &  \\
\addlinespace
  & 15 
    & $1.15\times10^{1}$ & $1.65\times10^{1}$ & $1.08\times10^{1}$ & \bm{$5.59\times10^{0}$}
    & $1.22\times10^{1}$ & $1.69\times10^{1}$ & $1.16\times10^{1}$ & \bm{$7.01\times10^{0}$}
    & 11,796,480 & 692,388,882 & 704,185,362 & 704,185,380 \\ 
  & 
    & ($\pm0.00$) & ($\pm1.65\times10^{1}$) & ($\pm1.79\times10^{-1}$) & (\bm{$\pm3.82\times10^{-2}$})
    & ($\pm0.00$) & ($\pm1.62\times10^{1}$) & ($\pm1.69\times10^{-1}$) & (\bm{$\pm3.63\times10^{-2}$})
    &  &  &  &  \\
\addlinespace
  & 20 
    & $9.78\times10^{0}$ & $1.94\times10^{1}$ & $9.13\times10^{0}$ & \bm{$4.54\times10^{0}$}
    & $1.08\times10^{1}$ & $1.97\times10^{1}$ & $1.02\times10^{1}$ & \bm{$6.45\times10^{0}$}
    & 15,728,640 & 693,699,607 & 709,428,247 & 709,428,270 \\ 
  & 
    & ($\pm0.00$) & ($\pm1.82\times10^{1}$) & ($\pm9.70\times10^{-2}$) & (\bm{$\pm3.73\times10^{-2}$})
    & ($\pm0.00$) & ($\pm1.78\times10^{1}$) & ($\pm9.29\times10^{-2}$) & (\bm{$\pm3.39\times10^{-2}$})
    &  &  &  &  \\
\bottomrule
\end{tabular}
}
\caption{Summary of training $L^2$ error, testing $L^2$ error, and number of parameters for the four models trained on data with varying resolutions $N$ and ranks $r$ pertaining to 3D HIT case.}
\label{tab:hit_table}
\end{table*}

\section{Training Time}\label{sec:train_time}
\begin{figure}[H]
\centering
\includegraphics[width=0.8\textwidth]{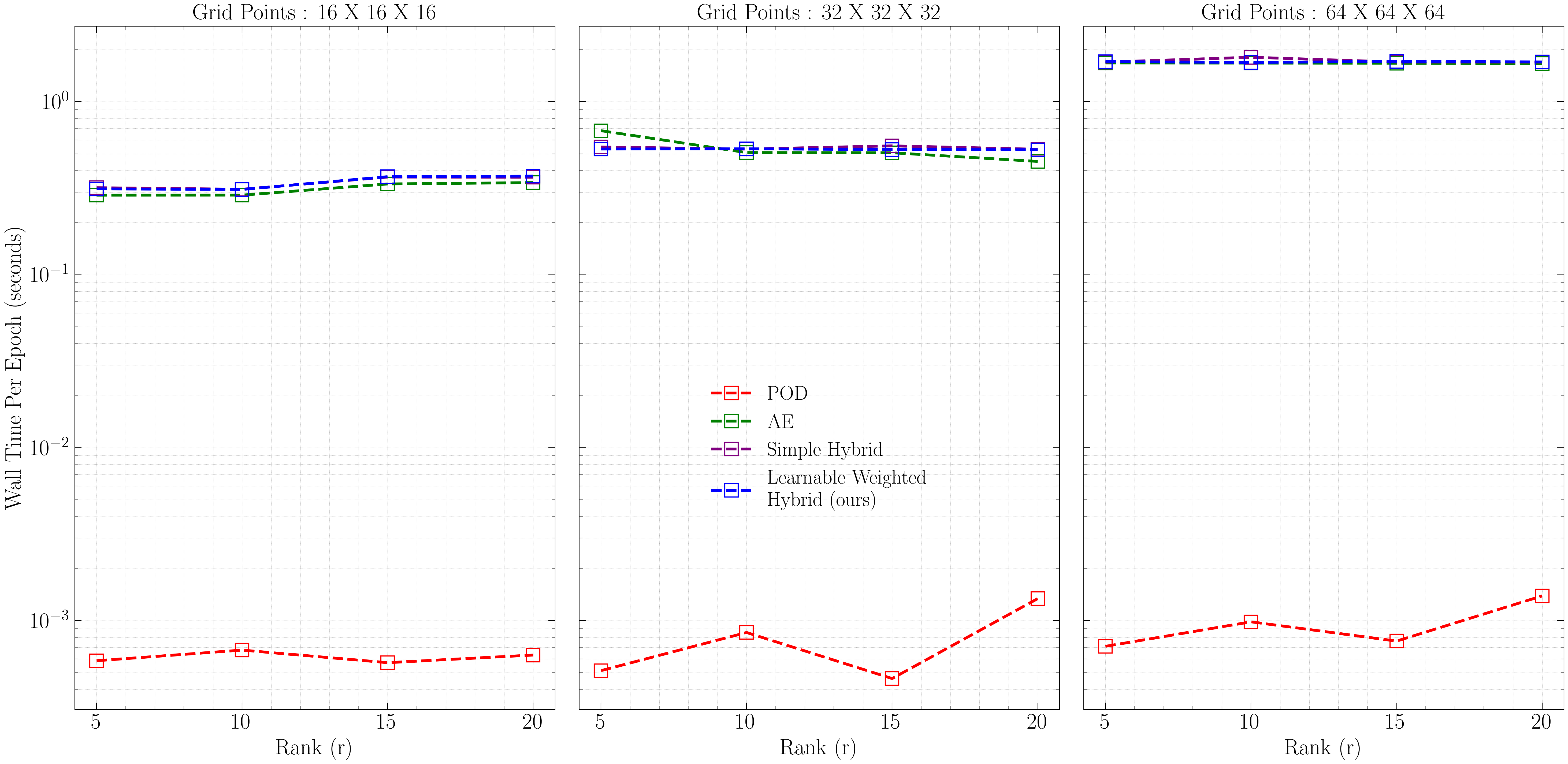}
%\rmfamily
\caption{Average wall time per epoch for the different methods pertaining to 3D HIT case. The comparable wall time across non-linear methods confirms that the learnable weighted approach does not introduce additional computational overhead.}
\label{fig:wall_time}
\end{figure}

\textcolor{black}{The wall time taken per epoch during training for each of the methods is shown in \Cref{fig:wall_time}. All the non-linear dimensionality reduction techniques have similar computational wall time per epoch during training indicating that there is no additional computational overhead incurred in training the proposed approach as compared to other techniques using deep learning models with similar number of parameters.}

\section{Comparison with $\beta$-VAE Architecture}\label{sec:beta_vae}
\begin{figure}[H]
\centering
\includegraphics[width=1.0\textwidth]{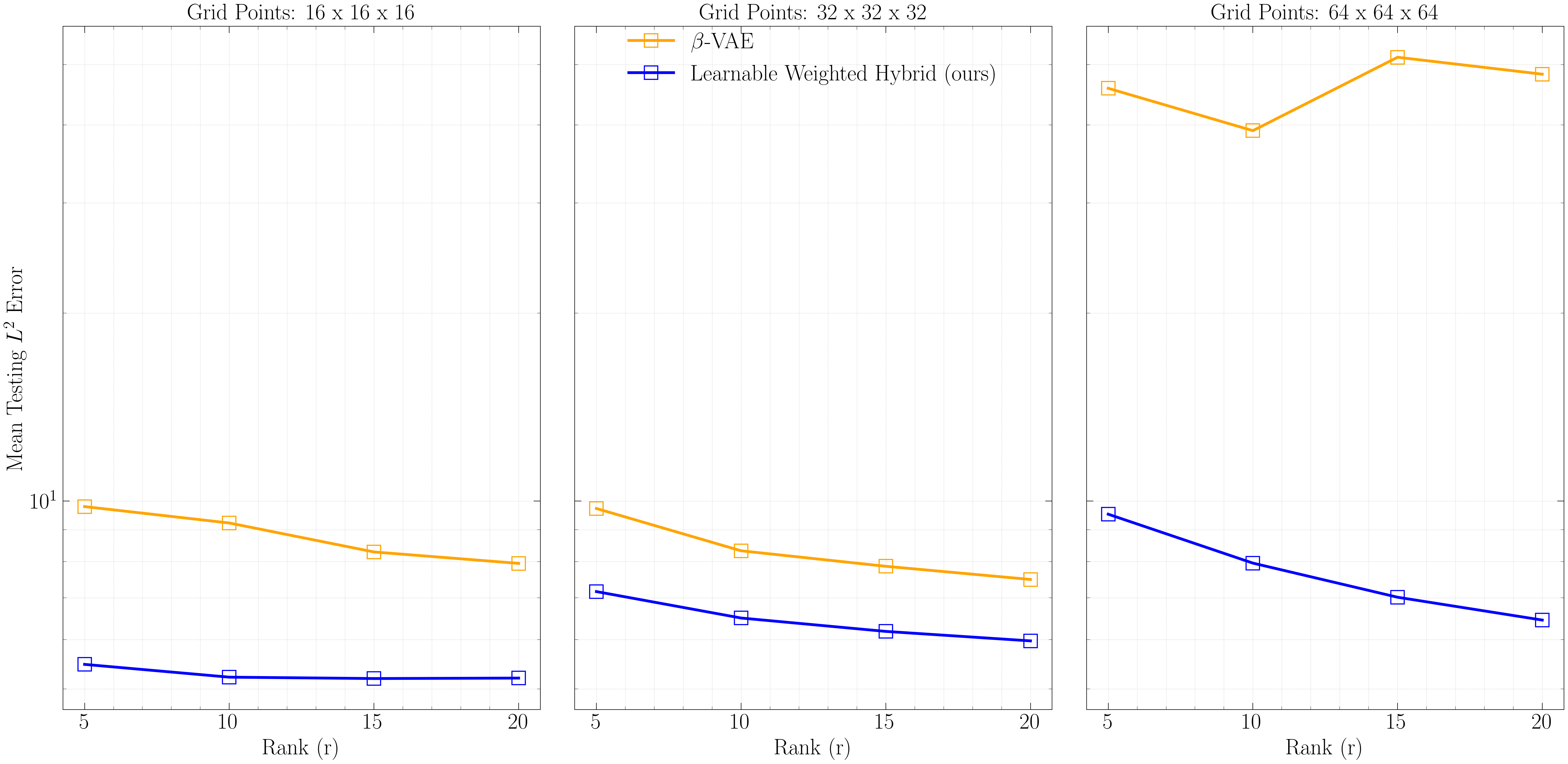}
%\rmfamily
\caption{Comparison of the Testing $L^2$ error between Learnable Weighted Hybrid (ours) and $\beta$ VAE for the 3D HIT case.}
\label{fig:betaae}
\end{figure}

\textcolor{black}{We compare the performance of our proposed approach with an existing architecture in literature \cite{solera2024beta} as shown in \Cref{fig:betaae}. The number of parameters is kept nearly the same for both the methods and trained for the same number of epochs and other learning settings. The proposed approach shows superior performance over the $\beta$-VAE for all the grid resolutions and latent space ranks. The performance of the $\beta$-VAE degrades significantly at higher resolutions, particularly at $64^3$, where notable training instabilities were also encountered.}   

\section{Relative Contribution of POD and NN}\label{sec:rel_contrib}

\begin{figure}[H]
\centering
\includegraphics[width=1.0\textwidth]{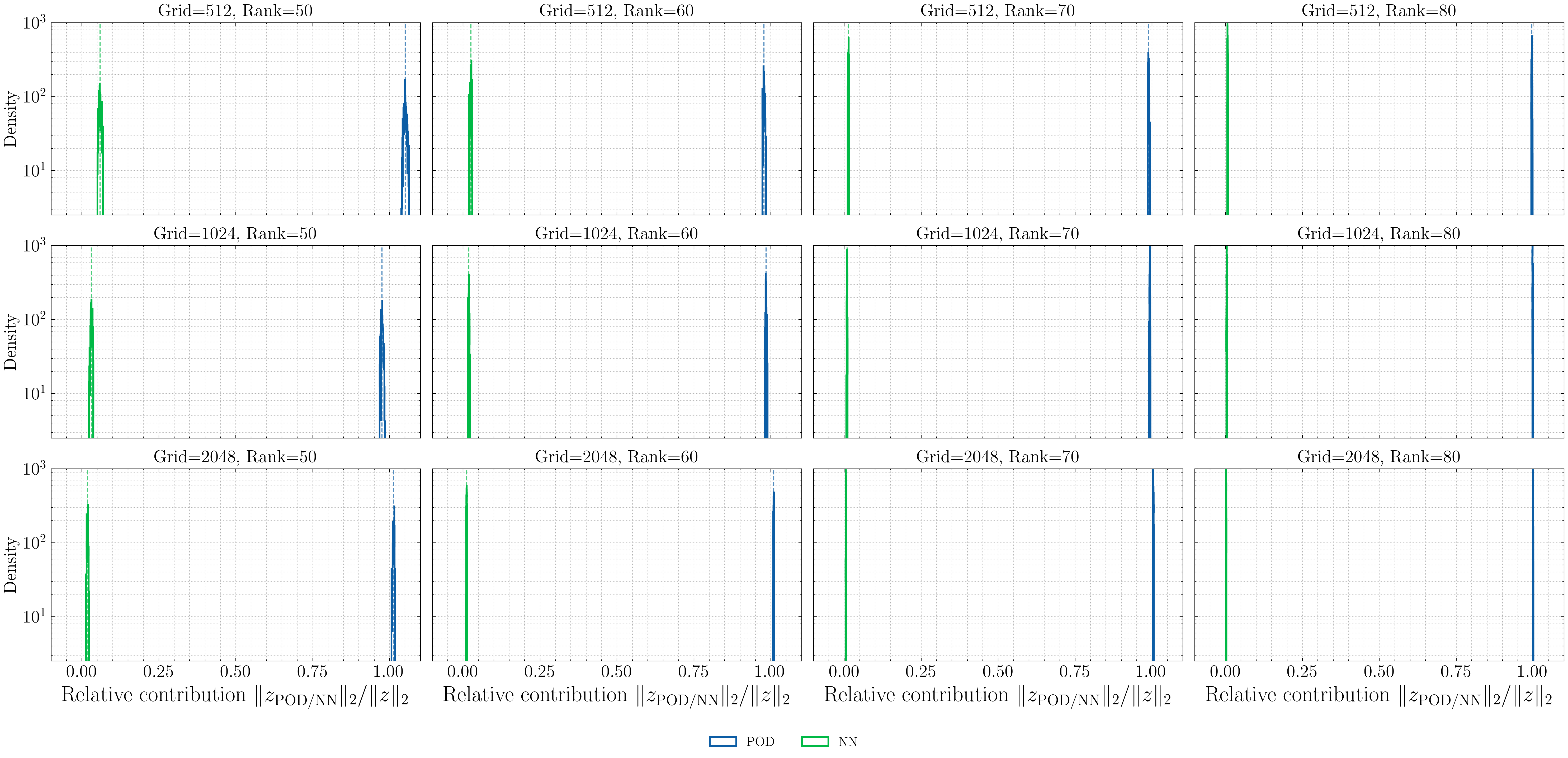}
%\rmfamily
\caption{Normalized contribution of POD and NN towards the latent representation for 1D KS case.}
\label{fig:z1_ks}
\end{figure}

\begin{figure}[H]
\centering
\includegraphics[width=1.0\textwidth]{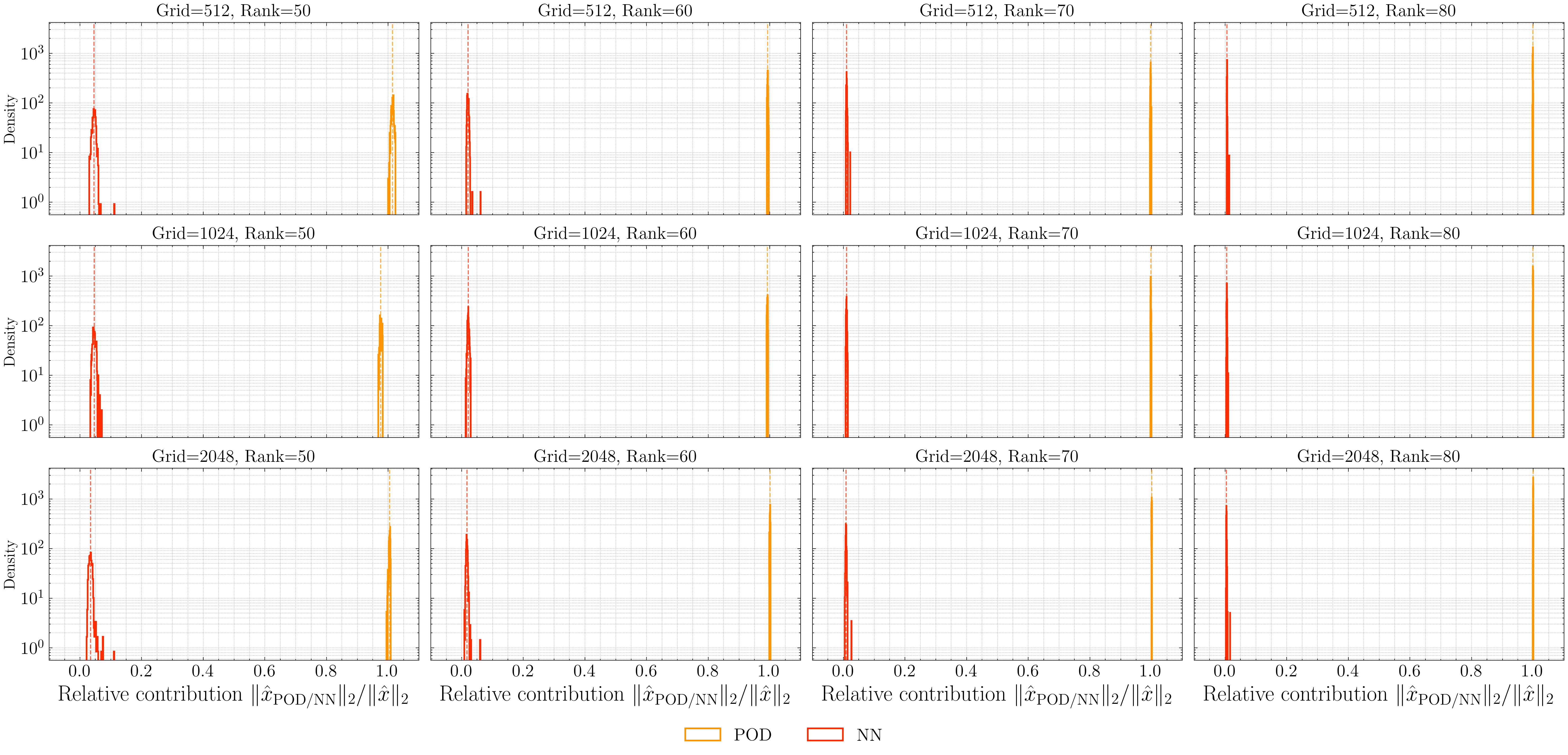}
%\rmfamily
\caption{Normalized contribution of POD and NN towards the reconstruction for 1D KS case.}
\label{fig:z2_ks}
\end{figure}

\begin{figure}[H]
\centering
\includegraphics[width=1.0\textwidth]{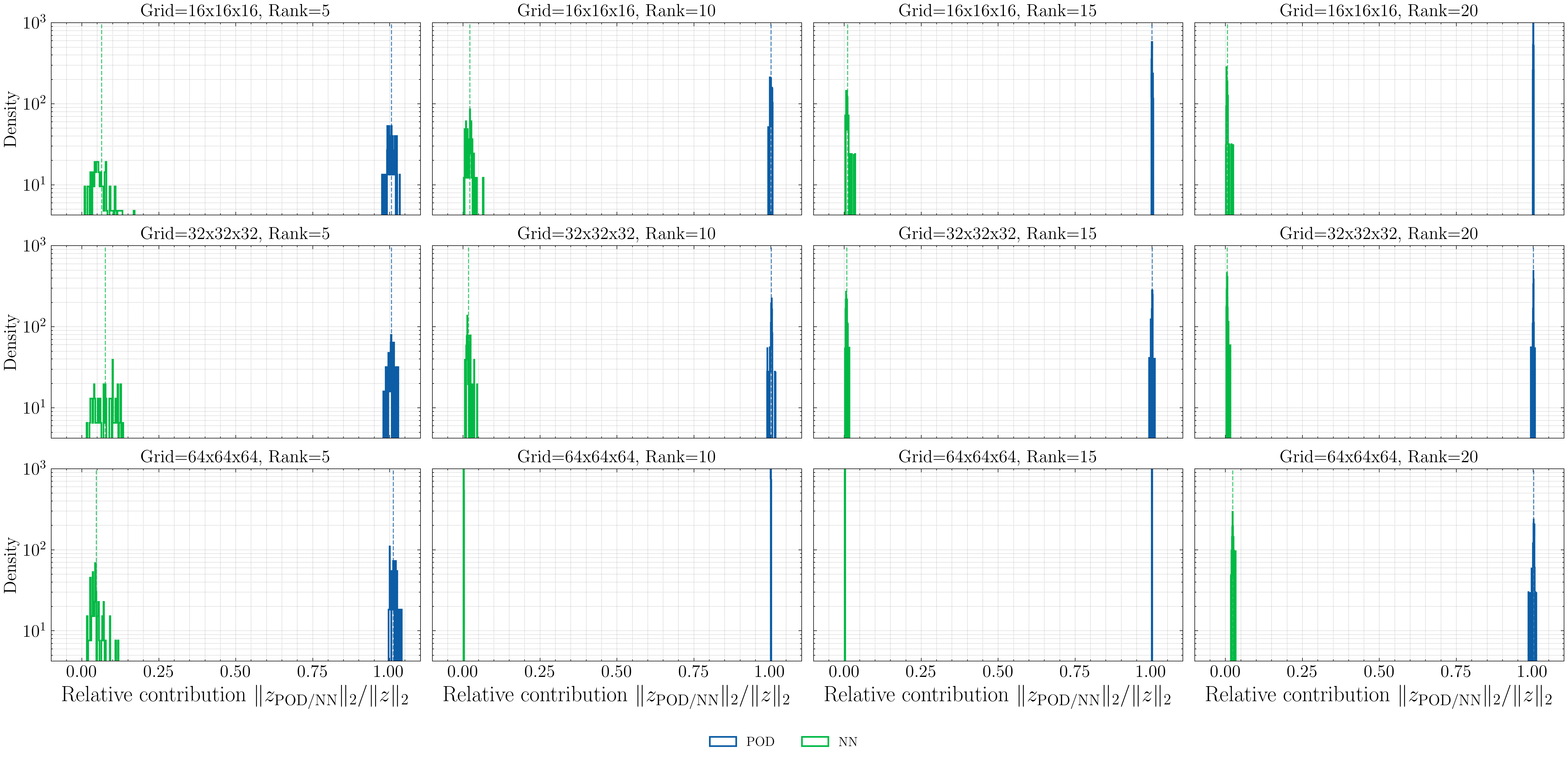}
%\rmfamily
\caption{Normalized contribution of POD and NN towards the latent representation for 3D HIT case.}
\label{fig:z1_hit}
\end{figure}

\begin{figure}[H]
\centering
\includegraphics[width=1.0\textwidth]{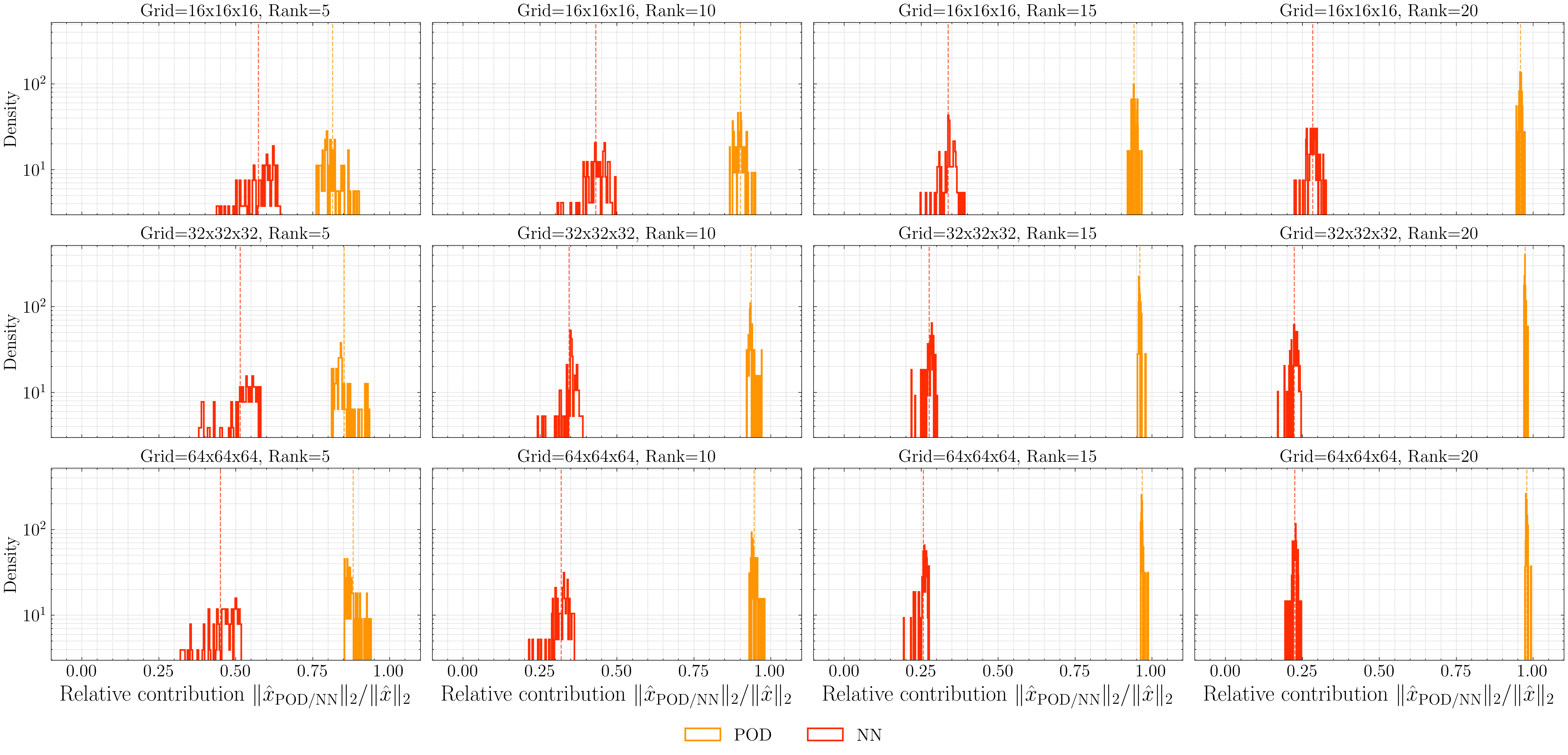}
%\rmfamily
\caption{Normalized contribution of POD and NN towards the reconstruction for 3D HIT case.}
\label{fig:z2_hit}
\end{figure}

\textcolor{black}{\Cref{fig:z1_ks} shows the distribution of normalized contribution by POD and NN towards the latent representation for various grid and ranks for the 1D KS case using the proposed approach. The contribution from the two towards the reconstruction is shown in \Cref{fig:z2_ks}. For this dataset, it can be seen that the contribution from POD dominates over NN in both latent representation and reconstruction. The same for the 3D HIT dataset is visualized in \Cref{fig:z1_hit} and \Cref{fig:z2_hit}. In this case, the reconstruction seems to have notable contribution from NN as compared to the 1D KS.} 

\section{Latent Representation Robustness}\label{sec:latentrep_robust}
\begin{figure}[!h]
\centering
\includegraphics[width=0.9\textwidth]{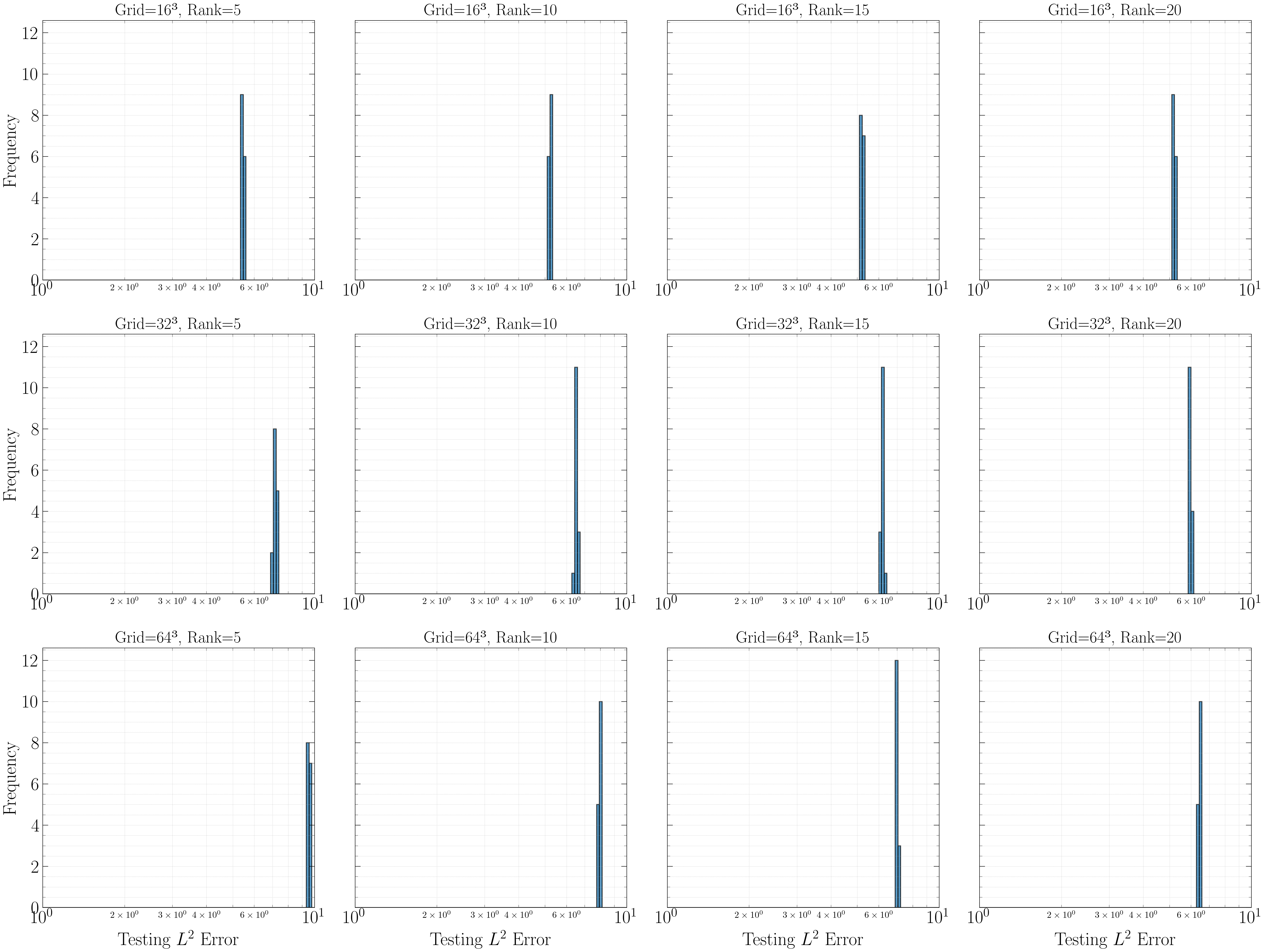}
%\rmfamily
\caption{Testing $L^2$ error distribution across various 15 different training initializations for the 3D HIT case using Learnable Weighted Hybrid (ours) approach. The error has little to no variance demonstrating the robustness of the proposed framework}
\label{fig:error_freq}
\end{figure}

\begin{figure}[!h]
\centering
\includegraphics[width=0.9\textwidth]{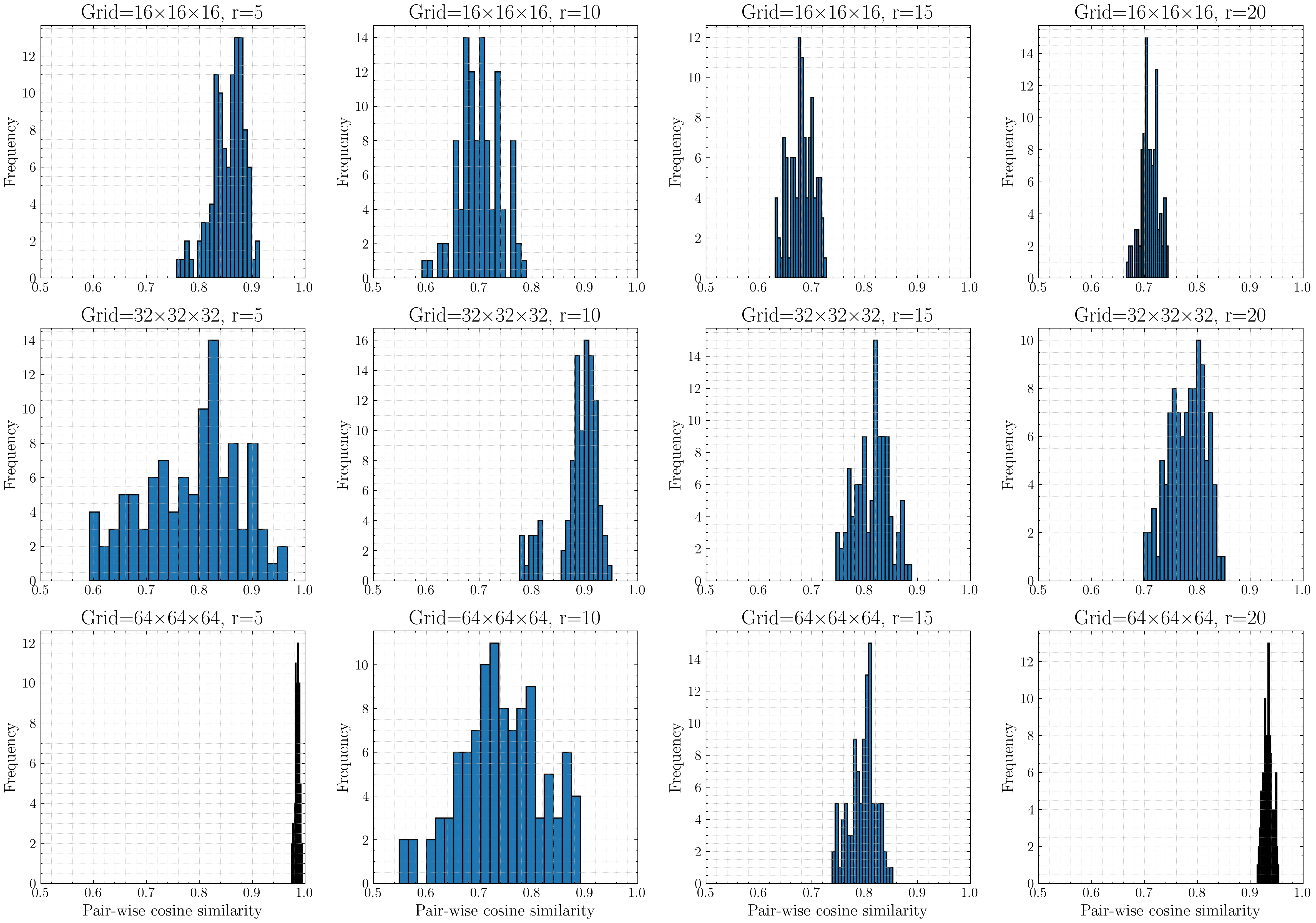}
%\rmfamily
\caption{Pair wise cosine similarity distribution in the latent representation across 15 different training initializations for the 3D HIT case using Learnable Weighted Hybrid (ours) approach. The similarity score range between 0.6 and 0.9 suggesting high degree of reproducibility in the latent representation.}
\label{fig:cosine}
\end{figure}

\textcolor{black}{We demonstrate the reproducibility and stability of the proposed approach in \Cref{fig:error_freq}, where the distribution of the testing $L_2$ error for 3D HIT case using Learnable Weighted Hybrid (ours) framework
is shown for various grid and subspace rank combinations. This distribution is computed over multiple training initializations. The consistently low variability in testing error highlights the robustness and reliability of our method with respect to random training initializations.  
The variability in the learned latent space representations for the 3D HIT dataset is illustrated in \Cref{fig:cosine}. We compute the pairwise cosine similarity between representations obtained from different training initializations.
The similarity scores vary between 0.6 and 0.9 for all grid resolutions and ranks, demonstrating reproducibility in latent representation.}    

\FloatBarrier

%%%%%%%%%% Insert bibliography here %%%%%%%%%%%%%%

\vskip2pc

%\noindent {\bf Please follow the coding for references as shown below.}

%\bibitem{1} Allwood JM, Cullen JM. 2011 \textit{Sustainable materials:  with both eyes open}.
%Cambridge, UK: UIT Cambridge. See \href{http://www.withbotheyesopen.com}{http://www.withbotheyesopen.com}.

%\bibitem{2}  MacKay DJC. 2008  \textit{Sustainable energy:  without the hot air}.
% Cambridge, UK: UIT Cambridge. See \href{http://www.withouthotair.com}{http://www.withouthotair.com}.

%\bibitem{3} Gallman PG. 2011  \textit{Green alternatives and national energy strategy: the facts
% behind the headlines}.  Baltimore,\ MD: Johns Hopkins University Press.

%\bibitem{4} MacKay DJC. 2013.  Solar energy in the context of energy use, energy transportation, and
% energy storage. \textit{Proc. R. Soc. A} \textbf{371}.
\bibliographystyle{RS} %%%% .BST file

\bibliography{panlab} %%%%% .Bib file

%\noindent If maintaining .bib file for references, then please use "RS.bst" to generate the references.

%\noindent Example:

\end{document}